\documentclass{article}
\usepackage{algorithm}
\usepackage{algorithmic}

\usepackage[final]{neurips_2025}

\usepackage[utf8]{inputenc} 
\usepackage[T1]{fontenc}    
\usepackage{hyperref}       
\usepackage{url}            
\usepackage{booktabs}       
\usepackage{amsfonts}       
\usepackage{graphicx}       
\usepackage{nicefrac}       
\usepackage{microtype}      
\usepackage{xcolor}         
\usepackage{multirow}
\usepackage{colortbl}
\usepackage{subfigure}
\usepackage{amsmath}
\usepackage{wrapfig}
\usepackage{enumitem}
\usepackage{amsthm}

\title{HyperGraphRAG: Retrieval-Augmented Generation via Hypergraph-Structured Knowledge Representation}

\author{
{\bf Haoran Luo\textsuperscript{\rm 1,2}, Haihong E\textsuperscript{\rm 1}\thanks{\ \ Corresponding author.}\ , Guanting Chen\textsuperscript{\rm 1}, Yandan Zheng\textsuperscript{\rm 2}, Xiaobao Wu\textsuperscript{\rm 2}, Yikai Guo\textsuperscript{\rm 3},} \\
{\bf Qika Lin\textsuperscript{\rm 4}, Yu Feng\textsuperscript{\rm 5}, Zemin Kuang\textsuperscript{\rm 6}, Meina Song\textsuperscript{\rm 1}, Yifan Zhu\textsuperscript{\rm 1}, Luu Anh Tuan\textsuperscript{\rm 2}} \\
         \textsuperscript{1}Beijing University of Posts and Telecommunications\ \ \textsuperscript{2}Nanyang Technological University \\
         \textsuperscript{3}Beijing Institute of Computer Technology and Application\ \ \textsuperscript{4}National University of Singapore \\ 
         \textsuperscript{5}China Mobile Research Institute\ \ \textsuperscript{6}Beijing Anzhen Hospital, Capital Medical University \\ 
         \texttt{haoran.luo@ieee.org, ehaihong@bupt.edu.cn, anhtuan.luu@ntu.edu.sg}
}

\begin{document}

\maketitle
\setcounter{footnote}{0}  
\begin{abstract}
Standard Retrieval-Augmented Generation (RAG) relies on chunk-based retrieval, whereas GraphRAG advances this approach by graph-based knowledge representation. However, existing graph-based RAG approaches are constrained by binary relations, as each edge in an ordinary graph connects only two entities, limiting their ability to represent the n-ary relations ($n \geq 2$) in real-world knowledge. In this work, we propose \textbf{HyperGraphRAG}, the first hypergraph-based RAG method that represents n-ary relational facts via hyperedges. HyperGraphRAG consists of a comprehensive pipeline, including knowledge hypergraph construction, retrieval, and generation. Experiments across medicine, agriculture, computer science, and law demonstrate that HyperGraphRAG outperforms both standard RAG and previous graph-based RAG methods in answer accuracy, retrieval efficiency, and generation quality. Our data and code are publicly available\footnote{\ \url{https://github.com/LHRLAB/HyperGraphRAG}}.
\end{abstract}

\section{Introduction}

\begin{wrapfigure}{r}{0.5\textwidth}
\vspace{-4.5mm}
\centering
\includegraphics[width=6.75cm]{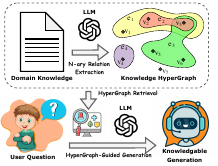}
\caption{An illustration of HyperGraphRAG. }
\label{F1}
\vspace{-2mm}
\end{wrapfigure}

Retrieval-Augmented Generation (RAG)~\citep{RAG,RAGsurvey} has advanced knowledge-intensive tasks by integrating knowledge retrieval with large language models (LLMs)~\citep{GPT4,LLMsurvey}, thereby enhancing factual awareness and generation accuracy. Standard RAG typically relies on chunk-based retrieval, segmenting documents into fixed-length text chunks retrieved via dense vector similarity, which overlooks the relationships between entities. Recently, GraphRAG~\citep{GraphRAG} has emerged as a promising direction that structures knowledge as a graph to capture inter-entity relations, with the potential to improve retrieval efficiency and knowledge-driven generation~\citep{KGLLMsurvey}.

\begin{figure}[t]
\centering
\includegraphics[width=13.2cm]{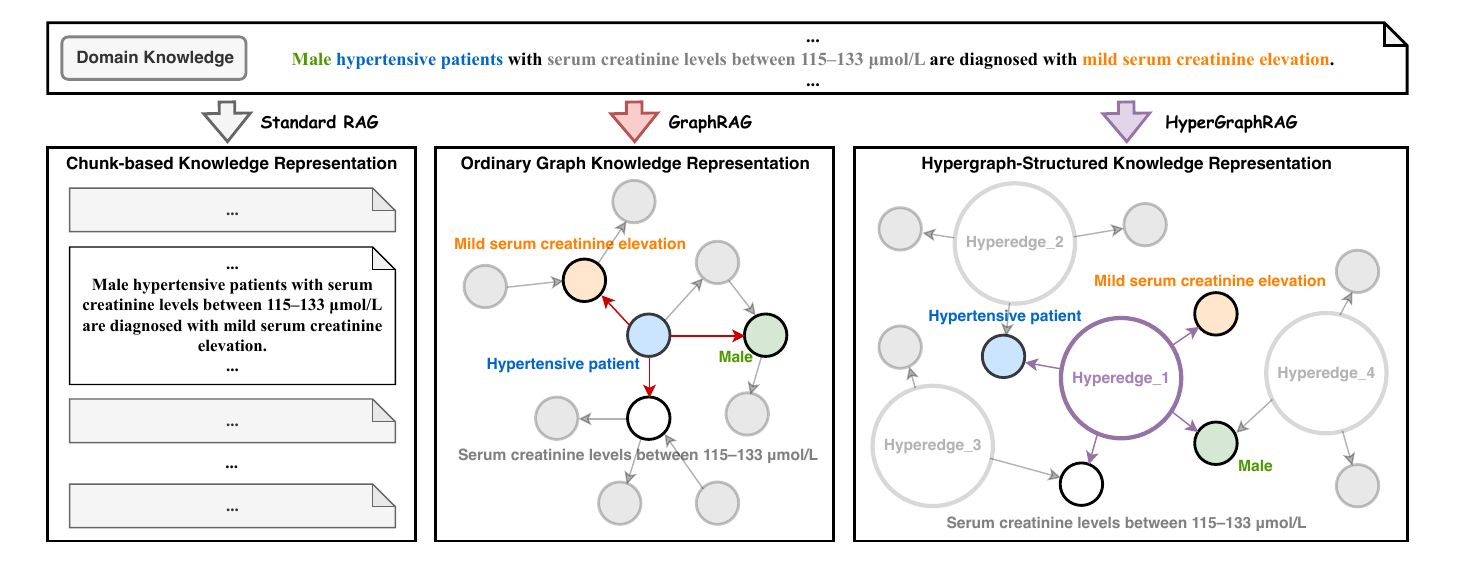}
\caption{Comparison of knowledge representation: standard RAG uses chunks as units, GraphRAG captures binary relations with graphs, and HyperGraphRAG models n-ary relations with hyperedges. }
\label{F2}
\end{figure}

However, since each edge in an ordinary graph connects only two entities, existing graph-based RAG approaches~\citep{GraphRAG,LightRAG,PathRAG,HippoRAG2} are all restricted to \textbf{binary relations}, making them insufficient for modeling the \textbf{n-ary relations among more than two entities} that are widespread in real-world domain knowledge~\citep{m-TransH}. For example, in the medical domain, as illustrated in Figure~\ref{F2}, representing the fact that ``\textit{Male hypertensive patients with serum creatinine levels between 115–133 $\mu$mol/L are diagnosed with mild serum creatinine elevation}'' requires decomposing it into several binary relational triples, such as \textit{Gender}:(\textit{Hypertensive patient}, \textit{Male}) and \textit{Diagnosed\_with}:(\textit{Hypertensive patient}, \textit{Mild serum creatinine elevation}), leading to representation sparsity during conversion process.

To address these limitations, we propose \textbf{HyperGraphRAG}, as illustrated in Figure~\ref{F1}, a novel graph-based RAG method built upon \textbf{hypergraph-structured knowledge representation}. In contrast to prior graph-based RAG methods constrained to binary relations, HyperGraphRAG leverages hyperedges to represent n-ary relational facts, where each hyperedge connects $n$ entities ($n\geq2$), e.g. \textit{Hyperedge}:(\textit{Hypertensive patient}, \textit{Male}, \textit{Serum creatinine levels between 115–133 $\mu$mol/L}, \textit{Mild serum creatinine elevation}), and each hyperedge is expressed through natural language descriptions. This design ensures knowledge completeness, structural expressiveness, and inferential capability, thereby providing more comprehensive support for knowledge-intensive applications.

Our proposed HyperGraphRAG is built upon three key steps. First, we propose a \textbf{knowledge hypergraph construction method}, leveraging LLM-based n-ary relation extraction to extract and structure multi-entity relationships. The resulting hypergraph is stored in a bipartite graph database, with separate vector databases for entities and hyperedges to facilitate efficient retrieval. Second, we develop a \textbf{hypergraph retrieval strategy} that employs vector similarity search to retrieve relevant entities and hyperedges, ensuring that the knowledge retrieved is both precise and contextually relevant. Lastly, we introduce a \textbf{hypergraph-guided generation mechanism}, which combines retrieved n-ary facts with traditional chunk-based RAG passages, thereby improving response quality.

To validate the effectiveness, we conduct experiments in multiple knowledge-intensive domains~\citep{LightRAG}, including medicine, agriculture, computer science, and law. Results demonstrate that HyperGraphRAG outperforms standard RAG and previous graph-based RAG methods in \textbf{answer accuracy}, \textbf{retrieval efficiency}, and \textbf{generation quality}, showcasing its strong potential for real-world applications.

\section{Related Work}
\textbf{Graph-based RAG. }
GraphRAG~\citep{GraphRAG} is the first graph-based RAG method that improves LLM generation via graph-based retrieval. Based on GraphRAG, several methods~\citep{MedGraphRAG,OG-RAG,KAG,MiniRAG,PIKE-RAG} focus on building graph-based RAG for different applications. LightRAG~\citep{LightRAG} enhances efficiency via graph indexing and updates. PathRAG~\citep{PathRAG} and HippoRAG2~\citep{HippoRAG2} refine retrieval with path pruning and Personalized PageRank. However, all rely on binary relations, limiting knowledge expressiveness. In this work, we propose HyperGraphRAG, the first graph-based RAG method via hypergraph-structured knowledge representation. We compare several existing methods with HyperGraphRAG in Table~\ref{comp}.

\textbf{Hypergraph Representation. }
Hypergraph-structured knowledge representation aims to overcome ordinary graph's limitations in modeling n-ary relations~\citep{Text2NKG}. Early methods~\citep{m-TransH,RAE,n-TuckER,HINGE} employ various embedding techniques to represent n-ary relational entities. Later methods~\citep{StarE,GRAN,HAHE} utilize GNN or attention to enhance embedding. However, existing methods mainly focus on link prediction, while hypergraphs also show potential for enhancing knowledge representation in graph-based RAG.

\begin{table}[t]
\caption{\label{comp}Comparison of knowledge construction and retrieval methods for NaiveGeneration, StandardRAG, partial GraphRAG baselines, and our proposed HyperGraphRAG, where $\mathcal{K}$ represents the overall constructed knowledge, and $K^*_q$ represents retrieved knowledge when given a user question $q$.}
\fontsize{8.4pt}{8.4pt}\selectfont
\centering
\resizebox{\linewidth}{!}{
\begin{tabular}{p{2.2cm}|p{5cm}|p{6.5cm}}
\toprule
\textbf{Method} & \textbf{Knowledge Construction} & \textbf{Knowledge Retrieval} \\
\midrule
NaiveGeneration & $\mathcal{K} = \emptyset$. & $K^*_q = \emptyset$.  \\

StandardRAG & $\mathcal{K} = \{c_i\}_{i=1}^N$, where $c_i$ is a chunk. & $K^*_q = K_{\text{chunk}} = \mathrm{Top}_k\{\,c \in \mathcal{K} \mid \mathrm{sim}(h_q,h_c)\,\}$ \\\midrule

GraphRAG~\citep{GraphRAG} & $\mathcal{K} = S=\{s_g \mid g \in \text{Community}(G)\}$,\newline where $S$ is the community summary set. & $K^*_q = \text{Detect}\{s_g \in S \mid q\}$,\newline where detected community summaries are retrieved. \\

LightRAG~\citep{LightRAG} & $\mathcal{K} = G=(V,E)$,\newline where $V$ \& $E$ are entity \& relation sets. & $K^*_q = \mathcal{F}\{v \in V, e \in E \mid q\} \cup K_{\text{chunk}}$,\newline where entities \& relations are retrieved with chunks.\\

PathRAG~\citep{PathRAG} & $\mathcal{K} = G=(V,E)$,\newline where $G$ is the same as LightRAG's. & $K^*_q = \text{Prune}\{p \in P_q \mid q\}$,\newline where relational paths are retrieved via pruning. \\

HippoRAG2~\citep{HippoRAG2} & $\mathcal{K} = G=(V\cup M,\;E)$,\newline where $V$ \& $M$ are phrase \& passage nodes. & $K^*_q = \text{PageRank}\{m \in M \mid q\}$,\newline where passages are retrieved via Personalized PageRank. \\\midrule 

HyperGraphRAG (ours) & $\mathcal{K} = G_H=(V,E_H)$,\newline where $G_H$ is structured as a hypergraph.  & 
$K^*_q = \mathcal{F}_n\{v \in V \mid q\} \cup \mathcal{F}_n\{e \in E_H \mid q\} \cup K_{\text{chunk}}$,\newline where n-ary relational facts are retrieved with chunks.\\
\bottomrule
\end{tabular}
}

\end{table}

\section{Preliminaries}
\textbf{Definition 1: RAG. }
Given a question $q$ and domain knowledge $K$, standard RAG first selects relevant document fragments $d$ from $K$ based on $q$, and then generates an answer $y$ based on $q$ and $d$. The probability model is formulated as:
\begin{equation}
P(y | q) = \sum_{d \in K} P(y | q, d) P(d | q, K).
\end{equation}
\textbf{Definition 2: Graph-based RAG. }
Graph-based RAG optimizes retrieval by representing knowledge as a graph structure $G = (V, E)$, where $V$ is the set of entities and $E$ is the set of relationships between entities. $G$ consists of facts represented as $F = (e, V_e) \in G$, where $e$ is the relation and $V_e$ is the entity set connected to $e$. Given a question $q$, the retrieval process is defined as:
\begin{equation}
P(y | q) = \sum_{F \in G} P(y | q, F) P(F | q, G).
\end{equation}
\textbf{Definition 3: Hypergraph. }
A hypergraph $G_H = (V, E_H)$~\citep{Hypergraph} is a generalized graph, where $V$ is the entity set, $E_H$ is the hyperedge set, and each hyperedge $e_H \in E_H$ connects 2 or more entities: 
\begin{equation}
V_{e_H} = (v_1, v_2, ..., v_n), \quad n \geq 2.
\end{equation}
Unlike ordinary graphs, where relationships are binary $V_e = (v_h, v_t)$, hypergraphs model n-ary relational facts $F_n = (e_H, V_{e_H}) \in G_H$.

\begin{figure*}[t]
\centering
\includegraphics[width=\linewidth]{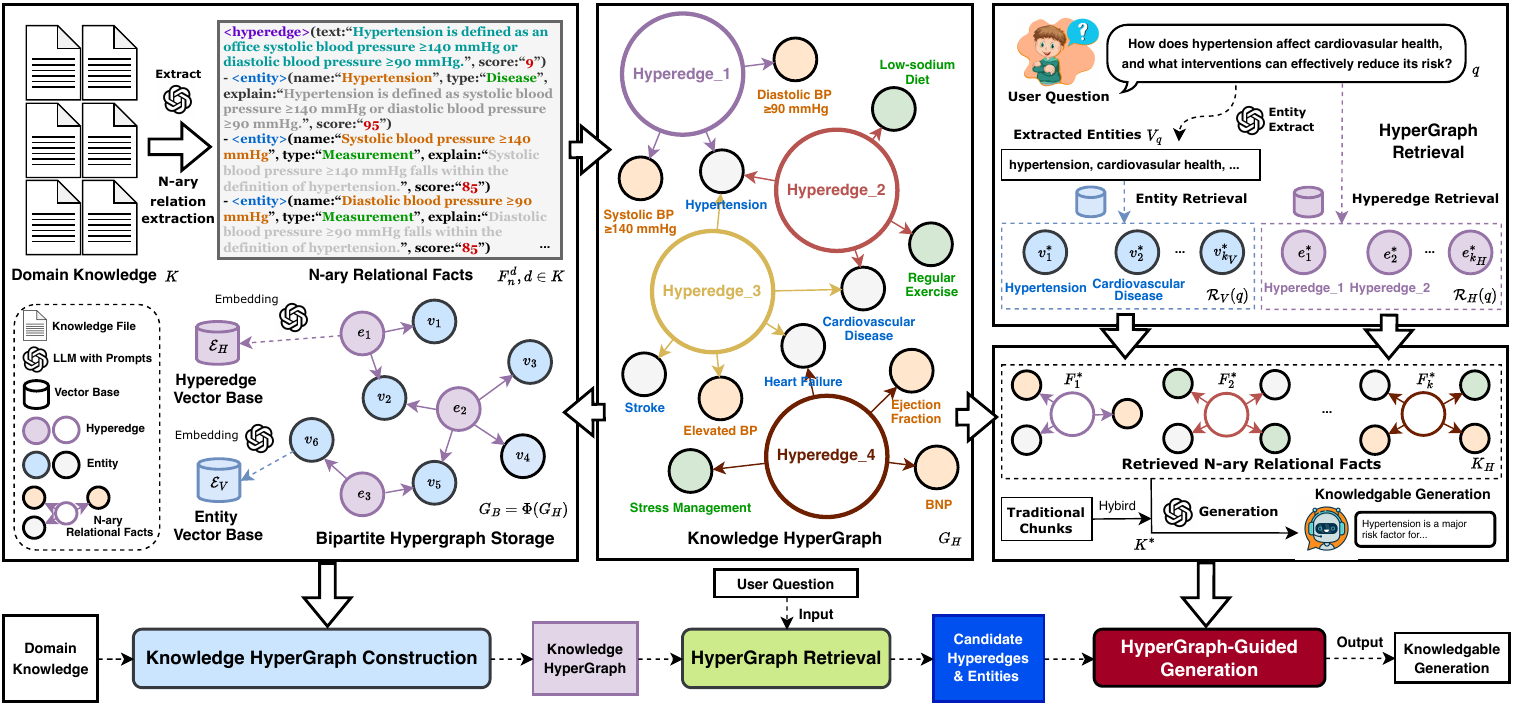}
\caption{An overview of HyperGraphRAG, which constructs a knowledge hypergraph from domain knowledge, retrieves n-ary facts based on user questions, and generates knowledgeable responses.}
\label{F3}
\end{figure*}

\section{Method: HyperGraphRAG}
In this section, we introduce the proposed HyperGraphRAG, as shown in Figure~\ref{F3}, including knowledge hypergraph construction, hypergraph retrieval strategy, and hypergraph-guided generation.

\subsection{Knowledge Hypergraph Construction}
To represent and store knowledge, we propose a knowledge hypergraph construction method that includes n-ary relational extraction, bipartite hypergraph storage, and vector representation storage.

\textbf{N-ary Relation Extraction.}
To construct the knowledge hypergraph $G_H$, our first step is to extract multiple n-ary relational facts $F_n$ from natural language documents $d \in K$. 
Unlike traditional hyper-relations~\citep{HINGE}, events~\citep{Text2Event}, or other n-ary relation models~\citep{Text2NKG}, in the era of LLMs, to preserve richer and more diverse n-ary relations among entities, we propose a new n-ary relation representation $F_n = (e_H, V_{e_H})$, utilizing \textbf{natural language descriptions}, instead of structured relations, to represent hyperedges $e_H$ among multiple entities $V_{e_H}$ as follows.
\begin{enumerate}[label=(\alph*), leftmargin=2em]
\item\textbf{Hyperedge:}
Given an input text $d$, it is parsed into several independent knowledge fragments, each treated as a hyperedge: $E_H^d = \{ e_1, e_2, ..., e_k \}$. Each hyperedge $e_i = (e^\text{text}_i, e^\text{score}_i)$ consists of two parts: a natural language description $e^\text{text}_i$, and a confidence score $e^\text{score}_i \in (0,10]$ indicating the association degree between $e_i$ and $d$.
\item\textbf{Entity:}
For each hyperedge $e_i$, entity recognition is performed to extract all contained entities: $V_{e_i} = \{ v_1, v_2, ..., v_n \}$, where $V_{e_i}$ is the entity set associated with $e_i$. Each entity $v_j = (v^\text{name}_j, v^\text{type}_j, v^\text{explain}_j, v^\text{score}_j)$ consists of four parts: entity name $v^\text{name}_j \subseteq e^\text{text}_i$, type $v^\text{type}_j$, explanation $v^\text{explain}_j$, and confidence score $v^\text{score}_j \in (0,100]$ indicating the extraction certainty. 
\end{enumerate}
Following this hypergraph-structured knowledge representation, we design an n-ary relation extraction prompt $p_{\text{ext}}$, detailed in Appendix~\ref{AppendixA1}, to enable the LLM $\pi$ to perform end-to-end knowledge fragment segmentation and entity recognition, thereby forming the n-ary relational fact set $F_n^d$:  
\begin{equation}
\label{E4}
F_n^d = \{ f_1, f_2, ..., f_k \} \sim \pi(F_n | p_{\text{ext}}, d),
\end{equation}
where each extracted n-ary relational fact $f_i = (e_i, V_{e_i})$ contains information about the corresponding hyperedge $e_i$ and its associated entity set $V_{e_i}$. We convert all documents $d \in K$ into hyperedges and entities using n-ary relation extraction, forming a complete knowledge hypergraph $G_H$.

\textbf{Proposition 1.} \textit{Hypergraph-structured knowledge representation is more comprehensive than binary.}\vspace{-2mm}
\begin{proof} 
We provide experimental results in Section~\ref{5.4} and proofs in Appendix~\ref{proof1}.
\end{proof}

\textbf{Bipartite Hypergraph Storage. }
After n-ary relation extraction, we store the constructed knowledge hypergraph $G_H$ in a graph database to support an efficient query. We adopt an ordinary graph database represented as a bipartite graph structure $G_B=(V_B, E_B)=\Phi(G_H)$, to store the knowledge hypergraph $G_H = (V,E_H)$, where $\Phi$ is a transformation function defined as:
\begin{equation}
\Phi:\ V_B = V \cup E_H,\ \ E_B = \{ (e_H, v) \mid e_H \in E_H, v \in V_{e_H} \}, 
\end{equation}
where $V_B$ is the set of nodes in $G_B$, formed by merging the entity set $V$ and the hyperedge set $E_H$ from $G_H$. The edge set $E_B$ captures the connections between each hyperedge $e_H \in E_H$ and its associated entities $v \in V_{e_H}$.

Based on $G_B$, we can efficiently query all entities associated with a hyperedge $e_H$ or query all hyperedges linked to a specific entity $v$, thereby benefiting the optimized query efficiency of an ordinary graph database, as well as preserving the complete hypergraph-structured knowledge representation.

Moreover, $G_B$ allows incremental updates through dynamically expansion: $G_B \leftarrow G_B \cup \Phi(G_H')$, where $G_H'$ represents newly added hypergraph information. The transformation of hyperedges and entities into the bipartite graph storage format enables seamless updates to the graph database.

\textbf{Proposition 2.} \textit{A bipartite graph can losslessly preserve and query a knowledge hypergraph.}\vspace{-2mm}
\begin{proof} 
We provide proofs in Appendix~\ref{proof2}.
\end{proof}

\textbf{Vector Representation Storage. }
To support efficient semantic retrieval, we embed hyperedges $e_H \in E_H$ and entities $v \in V$ using the same embedding model $f$, ensuring that the vector representation of hyperedges and entities is in the same vector space as questions. Let $\Psi$ be the vector function, then the vector representation storage for the knowledge hypergraph $G_H$ is defined as: $\Psi(G_H) = (\mathcal{E}_H, \mathcal{E}_V)$, where $\mathcal{E}_H$ is the vector base of hyperedges and $\mathcal{E}_V$ is the vector base of entities:
\begin{equation}
\Psi:\ \mathcal{E}_H = \{ \mathbf{h}_{e_H} \mid e_H \in E_H \},\ \ \mathcal{E}_V = \{ \mathbf{h}_v \mid v \in V \},
\end{equation}
where each hyperedge $e_H$ and entity $v$ in $G_H$ is embedded into their vector representations: $\mathbf{h}_{e_H} = f(e_H)$, and $\mathbf{h}_v = f(v)$, respectively.
  
\subsection{Hypergraph Retrieval Strategy}
After constructing and storing the hypergraph $G_H$, we design an efficient retrieval strategy to match user questions with relevant hyperedges and entities.

\textbf{Entity Retrieval. }
First, we extract key entities from the question $q$ to facilitate subsequent matching. We design an entity extraction prompt $p_{\text{q\_ext}}$, detailed in Appendix~\ref{AppendixA2}, along with the LLM $\pi$ to extract the entity set $V_q$:
\begin{equation}
\label{E7}
V_q \sim \pi(V | p_{\text{q\_ext}}, q).
\end{equation}
After extracting entities, we retrieve the most relevant entities from the entity set $V$ of the knowledge hypergraph $G_H$. We define the entity retrieval function $\mathcal{R}_V$, which retrieves the most relevant entities from $\mathcal{E}_V$ using cosine similarity:  
\begin{equation}
\mathcal{R}_V(q) = \operatorname*{argmax}^{k_V}_{v \in V} \left(\text{sim}(\mathbf{h}_{V_q}, \mathbf{h}_v) \odot v^{\text{score}}\right)_{>\tau_V},
\end{equation}
where $\mathbf{h}_{V_q} = f(V_q)$ is the concatenated text vector representation of the extracted entity set $V_q$, $\mathbf{h}_v \in \mathcal{E}_V$ is the vector representation of entity $v$, $\text{sim}(\cdot, \cdot)$ denotes the similarity function, $\odot$ represents element-wise multiplication between similarity and entity relevance score $v^{\text{score}}$ determining the final ranking score, $\tau_V$ is the threshold for the entity retrieval score, and $k_V$ is the limit on the number of retrieved entities.

\textbf{Hyperedge Retrieval. }
Moreover, to expand the retrieval scope and capture complete n-ary relations within the hyperedge set $E_H$ of the knowledge hypergraph $G_H$, we define the hyperedge retrieval function $\mathcal{R}_H$, which retrieves a set of hyperedges related to $q$:  
\begin{equation}
\mathcal{R}_H(q) = \operatorname*{argmax}^{k_H}_{e_H \in E_B} \left(\text{sim}(\mathbf{h}_q, \mathbf{h}_{e_H})  \odot e_H^{\text{score}}\right)_{>\tau_H},
\end{equation}
where $\mathbf{h}_q = f(q)$ is the text vector representation of $q$, $\mathbf{h}_{e_H} \in \mathcal{E}_H$ is the vector representation of the hyperedge $e_H$, $\odot$ represents element-wise multiplication between similarity and hyperedge relevance score $e_H^{\text{score}}$ determining the final ranking score, $\tau_H$ is the threshold for the hyperedge retrieval score, and $k_H$ limits the number of retrieved hyperedges.

\subsection{Hypergraph-Guided Generation} 
To fully utilize the structured knowledge in the hypergraph, we propose a Hypergraph-Guided Generation mechanism, which consists of hypergraph knowledge fusion and generation augmentation. 

\textbf{Hypergraph Knowledge Fusion. }
The primary goal of hypergraph knowledge fusion is to expand and reorganize the retrieved n-ary relational knowledge to form a comprehensive knowledge input. Since $q$ may only match partial entities or hyperedges, we further expand the retrieval scope. To obtain a complete set of n-ary relational facts, we design a bidirectional expansion strategy, that includes expanding hyperedges from retrieved entities and expanding entities from retrieved hyperedges.  

First, given the entity set retrieved from $q$, denoted as $\mathcal{R}_V(q) = \{ v_1, v_2, ..., v_{k_V} \}$, we retrieve all hyperedges in the knowledge hypergraph $G_H$ that connect these entities:
\begin{equation}
\mathcal{F}_V^* = \bigcup_{v_i \in \mathcal{R}_V(q)} \{ (e_H, V_{e_H}) \mid v_i \in V_{e_H}, e_H \in E_H \}.
\end{equation}  
Next, we expand the set of entities connected to the retrieved hyperedges $\mathcal{R}_H(q) = \{ e_1, e_2, ..., e_{k_H} \}$:  
\begin{equation}
\mathcal{F}_H^* = \bigcup_{e_i \in \mathcal{R}_H(q)} \{ (e_i, V_{e_i}) \mid V_{e_i} \subseteq V \}
\end{equation}  
Finally, we merge the expanded hyperedge set $\mathcal{F}_V^*$ with the expanded entity set $\mathcal{F}_H^*$ to form a complete retrieved n-ary relational fact set $K_H = \mathcal{F}_V^* \cup \mathcal{F}_H^*$. This set contains all necessary n-ary relational knowledge for reasoning and generation, ensuring a comprehensive input for the LLM.

\textbf{Generation Augmentation. }
Following hypergraph knowledge fusion, we augment the generation strategy to improve the accuracy and readability of the responses. We adopt a hybrid RAG fusion mechanism, combining hypergraph knowledge $K_H$ with retrieved chunk-based text fragments $K_{\text{chunk}}$ to form the final knowledge input.
We define the final knowledge input $K^* = K_H \cup K_{\text{chunk}}$, where $K_{\text{chunk}}$ consists of chunk-based text fragments retrieved using traditional RAG.

Finally, we use a retrieval-augmented generation prompt $p_{\text{gen}}$, detailed in Appendix~\ref{AppendixA3}, that combines hypergraph knowledge $K^*$ and the user question $q$ as input to LLM $\pi$ to generate final response $y^*$:
\begin{equation}
\label{E12}
y^* \sim \pi(y|p_{\text{gen}},K^*,q).
\end{equation} 
\textbf{Proposition 3.} \textit{Retrieving knowledge on a knowledge hypergraph improves retrieval efficiency compared to methods based on ordinary binary graphs, leading to gains in generation quality.}\vspace{-2mm}
\begin{proof} 
We provide experimental results in Sections~\ref{5.5} and \ref{5.6} and proofs in Appendix~\ref{proof3}.
\end{proof}

\section{Experiments}
This section presents the experimental setup, main results, and analysis. We answer the following research questions (RQs):
\textbf{RQ1: }Does HyperGraphRAG outperform other methods?
\textbf{RQ2: }Does the main component of HyperGraphRAG work?
\textbf{RQ3: }How effective is the knowledge hypergraph constructed by HyperGraphRAG across various domains?
\textbf{RQ4: }Could the hypergraph retrieval strategy improve retrieval efficiency?
\textbf{RQ5: }How effective is the generation quality of HyperGraphRAG?
\textbf{RQ6: }How are the time and cost of HyperGraphRAG in construction and generation phases?

\subsection{Experimental Setup}

\textbf{Datasets. }
To evaluate the performance of HyperGraphRAG across multiple domains, we select four knowledge contexts from UltraDomain~\citep{UltraDomain}, as used in LightRAG~\citep{LightRAG}: \textbf{Agriculture}, Computer Science (\textbf{CS}), \textbf{Legal}, and a mixed domain (\textbf{Mix}). In addition, we include the latest international hypertension guidelines~\citep{Hypertension} as the foundational knowledge for the \textbf{Medicine} domain. For each of the five domains, we sample knowledge fragments one, two, and three hops away to construct questions with ground-truth answers verified by human annotators. We then categorize the questions into \textbf{Binary Source} and \textbf{N-ary Source}, based on whether the sampled knowledge of the question contains facts among $n$ entities ($n>2$). More details can be found in Appendix~\ref{AppendixA}.  

\textbf{Baselines. }
We compare HyperGraphRAG against six publicly available baseline methods: \textbf{NaiveGeneration}~\citep{GPT4}, which directly generates responses using LLM; \textbf{StandardRAG}~\citep{RAGsurvey}, a traditional chunk-based RAG approach; \textbf{GraphRAG}~\citep{GraphRAG}, \textbf{LightRAG}~\citep{LightRAG}, \textbf{PathRAG}~\citep{PathRAG}, and \textbf{HippoRAG2}~\citep{HippoRAG2}, which are four selected available graph-based RAG methods described in Table~\ref{comp}. To ensure fairness, we use the same generation prompt, which can be found in Appendix~\ref{AppendixE}.

\textbf{Evaluation Metrics. }
We evaluate the answer accuracy, retrieval efficiency, and generation quality of HyperGraphRAG and its baselines using 3 key metrics: \textbf{F1}, Retrieval Similarity (\textbf{R-S}), and Generation Evaluation (\textbf{G-E}). F1 measures word-level similarity between the generated answer and the ground-truth answer, following FlashRAG~\citep{FlashRAG}. R-S assesses the semantic similarity between the retrieved knowledge and the ground-truth knowledge used to construct the question, in line with RAGAS~\citep{RAGAS}. G-E, inspired by HelloBench~\citep{HelloBench}, is a metric that uses LLM-as-a-judge to evaluate generation quality in 7 dimensions and reports the average score. Details are provided in Appendix~\ref{AppendixE}.

\textbf{Implementation Details. }
We use OpenAI's \texttt{GPT-4o-mini} for extraction and generation, and \texttt{text-embedding-3-small} for vector. During retrieval, we set the following parameters: entity retrieval $k_V = 60$, $\tau_V = 50$; hyperedge retrieval $k_H = 60$, $\tau_H = 5$; and chunk retrieval $k_C = 5$, $\tau_C = 0.5$. All experiments were conducted on a server with an 80-core CPU and 512GB RAM.

\begin{table}[t]
\caption{\label{T1}
Performance comparison across different domains. \textbf{Bold} indicates the best performance.}
\centering
\fontsize{7.3pt}{8.3pt}\selectfont
\setlength{\tabcolsep}{0.8mm}{
\begin{tabular}{lccc|ccc|ccc|ccc|ccc}
\toprule
\multirow{2.5}{*}{\textbf{Method}} & \multicolumn{3}{c}{\textbf{Medicine}} & \multicolumn{3}{c}{\textbf{Agriculture}} & \multicolumn{3}{c}{\textbf{CS}} & \multicolumn{3}{c}{\textbf{Legal}} & \multicolumn{3}{c}{\textbf{Mix}} \\
\cmidrule(lr){2-4} \cmidrule(lr){5-7} \cmidrule(lr){8-10} \cmidrule(lr){11-13} \cmidrule(lr){14-16}
& \textbf{F1} & \textbf{R-S} & \textbf{G-E} & \textbf{F1} & \textbf{R-S} & \textbf{G-E} & \textbf{F1} & \textbf{R-S} & \textbf{G-E} & \textbf{F1} & \textbf{R-S} & \textbf{G-E} & \textbf{F1} & \textbf{R-S} & \textbf{G-E} \\
\midrule
\multicolumn{16}{c}{\textbf{\textit{Binary Source}}} \\
\midrule
\rowcolor{gray!10} NaiveGeneration & 12.63 & 0.00 & 44.70 & 11.71 & 0.00 & 45.76 & 18.93 & 0.00 & 48.79 & 22.91 & 0.00 & 50.00 & 18.58 & 0.00 & 46.14 \\
\rowcolor{gray!10} StandardRAG & 26.87 & 61.08 & 56.24 & 28.31 & 42.69 & 57.58 & 28.87 & 49.44 & 57.10 & 37.19 & 52.21 & 59.85 & 47.57 & 46.79 & 67.42 \\
GraphRAG & 17.13 & 54.56 & 48.19 & 20.67 & 40.90 & 52.41 & 23.75 & 37.65 & 53.17 & 31.09 & 34.26 & 54.62 & 23.62 & 25.01 & 48.12 \\
LightRAG & 12.16 & 52.38 & 44.15 & 17.70 & 41.24 & 50.32 & 22.59 & 41.86 & 51.62 & 33.63 & 45.54 & 56.42 & 29.98 & 34.22 & 54.50 \\
PathRAG & 14.74 & 52.30 & 45.36 & 21.97 & 42.21 & 53.13 & 25.28 & 41.49 & 53.28 & 32.32 & 43.60 & 55.45 & 40.87 & 33.36 & 60.75 \\
HippoRAG2 & 21.12 & 57.50 & 51.08 & 12.60 & 16.85 & 44.56 & 16.94 & 21.05 & 47.29 & 20.10 & 34.13 & 46.77 & 21.10 & 18.34 & 45.83 \\
\rowcolor{blue!10} HyperGraphRAG (ours) & \textbf{36.45} & \textbf{69.91} & \textbf{60.65} & \textbf{34.80} & \textbf{61.97} & \textbf{59.99} & \textbf{31.60} & \textbf{60.94} & \textbf{57.54} & \textbf{44.42} & \textbf{60.87} & \textbf{63.53} & \textbf{51.51} & \textbf{67.34} & \textbf{68.76} \\
\midrule
\multicolumn{16}{c}{\textbf{\textit{N-ary Source}}} \\
\midrule
\rowcolor{gray!10} NaiveGeneration & 13.15 & 0.00 & 41.83 & 13.78 & 0.00 & 47.93 & 18.37 & 0.00 & 48.94 & 20.37 & 0.00 & 48.09 & 15.29 & 0.00 & 45.16 \\
\rowcolor{gray!10} StandardRAG & 28.93 & 64.06 & 55.08 & 26.55 & 48.93 & 56.62 & 28.99 & 47.35 & 56.69 & 37.50 & 51.16 & 60.09 & 38.83 & 47.73 & 61.82 \\
GraphRAG & 18.07 & 57.22 & 47.09 & 21.90 & 41.27 & 53.49 & 22.90 & 39.97 & 53.76 & 29.12 & 34.11 & 53.76 & 14.93 & 24.32 & 42.32 \\
LightRAG & 13.43 & 54.67 & 41.86 & 18.78 & 42.44 & 50.92 & 22.85 & 41.19 & 52.20 & 29.64 & 44.47 & 54.65 & 24.08 & 33.22 & 50.83 \\
PathRAG & 15.14 & 54.08 & 42.77 & 20.64 & 42.53 & 51.83 & 28.18 & 42.29 & 54.97 & 30.27 & 44.47 & 55.26 & 33.27 & 34.11 & 57.47 \\
HippoRAG2 & 21.56 & 61.54 & 48.06 & 12.66 & 20.32 & 45.14 & 17.75 & 26.92 & 48.44 & 16.95 & 34.72 & 45.09 & 21.95 & 18.49 & 46.87 \\
\rowcolor{blue!10} HyperGraphRAG (ours) & \textbf{34.26} & \textbf{70.48} & \textbf{58.06} & \textbf{32.98} & \textbf{62.58} & \textbf{59.59} & \textbf{31.00} & \textbf{59.25} & \textbf{58.35} & \textbf{43.20} & \textbf{60.07} & \textbf{63.70} & \textbf{45.91} & \textbf{69.09} & \textbf{65.04} \\
\midrule
\multicolumn{16}{c}{\textbf{\textit{Overall}}} \\
\midrule
\rowcolor{gray!10} NaiveGeneration & 12.89 & 0.00 & 43.27 & 12.74 & 0.00 & 46.85 & 18.65 & 0.00 & 48.87 & 21.64 & 0.00 & 49.05 & 16.93 & 0.00 & 45.65 \\
\rowcolor{gray!10} StandardRAG & 27.90 & 62.57 & 55.66 & 27.43 & 45.81 & 57.10 & 28.93 & 48.40 & 56.89 & 37.34 & 51.68 & 59.97 & 43.20 & 47.26 & 64.62 \\
GraphRAG & 17.60 & 55.89 & 47.64 & 21.28 & 41.08 & 52.95 & 23.33 & 38.81 & 53.47 & 30.11 & 34.18 & 54.19 & 19.27 & 24.67 & 45.22 \\
LightRAG & 12.79 & 53.52 & 43.00 & 18.24 & 41.84 & 50.62 & 22.72 & 41.53 & 51.91 & 31.64 & 45.00 & 55.53 & 27.03 & 33.72 & 52.67 \\
PathRAG & 14.94 & 53.19 & 44.06 & 21.30 & 42.37 & 52.48 & 26.73 & 41.89 & 54.13 & 31.29 & 44.03 & 55.36 & 37.07 & 33.73 & 59.11 \\
HippoRAG2 & 21.34 & 59.52 & 49.57 & 12.63 & 18.58 & 44.85 & 17.34 & 23.99 & 47.87 & 18.53 & 34.42 & 45.93 & 21.53 & 18.42 & 46.35 \\
\rowcolor{blue!10} HyperGraphRAG (ours) & \textbf{35.35} & \textbf{70.19} & \textbf{59.35} & \textbf{33.89} & \textbf{62.27} & \textbf{59.79} & \textbf{31.30} & \textbf{60.09} & \textbf{57.94} & \textbf{43.81} & \textbf{60.47} & \textbf{63.61} & \textbf{48.71} & \textbf{68.21} & \textbf{66.90} \\
\bottomrule
\end{tabular}}
\vspace{-0.1mm}
\end{table}

\subsection{Main Results (RQ1)}
To evaluate the effectiveness of HyperGraphRAG, we compare its performance with various baselines across multiple domains. The results are shown in Table~\ref{T1}.

\textbf{Overall Comparison Across Methods. }
HyperGraphRAG consistently outperforms all baselines across F1, R-S, and G-E metrics. Compared to StandardRAG, it achieves gains of +7.45 (F1), +7.62 (R-S), and +3.69 (G-E). Interestingly, existing graph-based RAG baselines often underperform StandardRAG, as their reliance on binary relational graphs causes knowledge fragmentation, sparsified retrieval, and incomplete context reconstruction during generation.

\textbf{Comparison Across Source Types. }
HyperGraphRAG maintains strong gains under both Binary and N-ary settings. For Binary Source, it improves F1, R-S, and G-E by +8.6, +8.8, and +4.4; for N-ary Source, the improvements are +5.3, +6.4, and +2.9, confirming its robustness.

\textbf{Comparison Across Domains. }
Performance gains are consistent across domains, with the largest improvements in Medicine and Legal (over +7 F1), and stable advantages in Agriculture and CS. HyperGraphRAG adapts well to both highly structured and more general knowledge tasks.


\subsection{Ablation Study (RQ2)}
As shown in Figure~\ref{F4}, we conduct an ablation study in the Medicine domain by removing entity retrieval (w/o ER), hyperedge retrieval (w/o HR), and their combination (w/o ER \& HR). We also remove chunk retrieval fusion (w/o CR), and all modules (w/o ER \& HR \& CR):

\begin{wrapfigure}{r}{0.5\textwidth}
\vspace{-1.5mm}
\centering
\includegraphics[width=6.9cm]{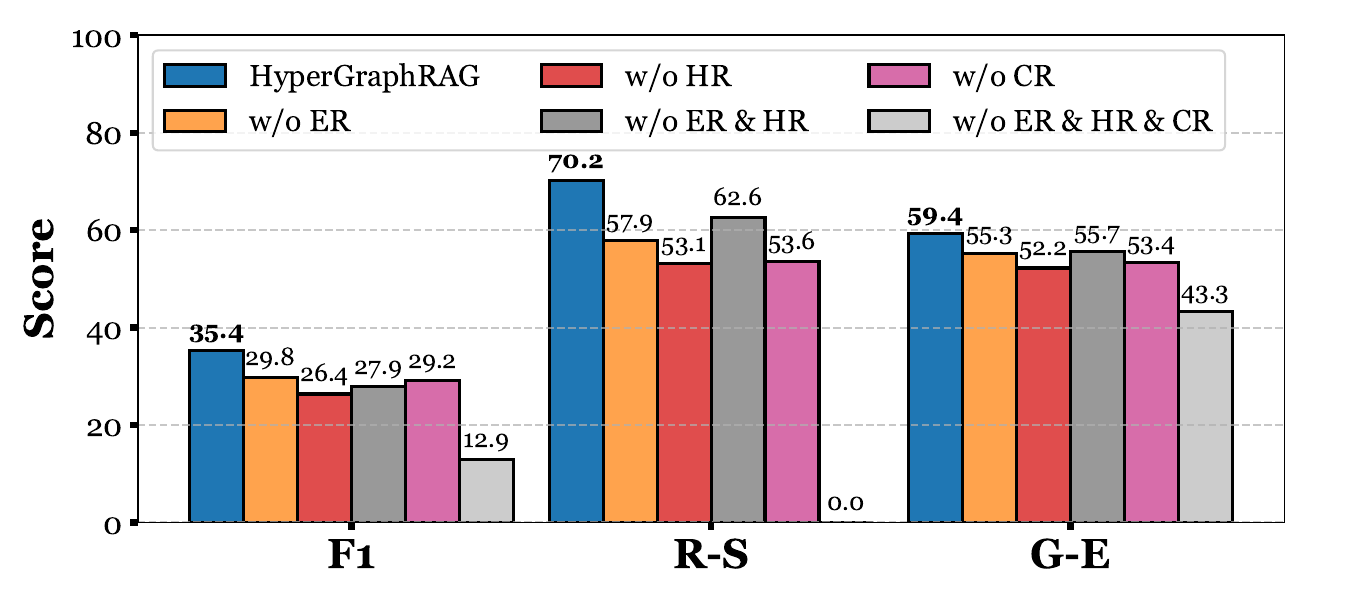}
\caption{Results of the ablation study. }
\label{F4}
\vspace{-2mm}
\end{wrapfigure}

\textbf{Impact of Entity Retrieval (ER). }
ER is critical for precise retrieval by anchoring key concepts. Without ER, F1 falls from 35.4 to 29.8, underscoring its importance in selecting relevant entities for accurate generation.

\textbf{Impact of Hyperedge Retrieval (HR). }
HR captures n-ary, multi-entity facts necessary for complex reasoning. Removing HR drops F1 from 35.4 to 26.4, highlighting its unique role beyond mere entity retrieval.

\textbf{Impact of Chunk Retrieval Fusion (CR). }
CR enhances retrieval by integrating unstructured text with hypergraph data. Excluding CR reduces F1 from 35.4 to 29.2, demonstrating that the fusion leads to more complete and fluent generation.

\subsection{Analysis of Hypergraph-structured Knowledge Representation (RQ3)}
\label{5.4}
As shown in Figure~\ref{F5}, we assess HyperGraphRAG's knowledge representation across 5 domains:
\begin{figure}[h]
\vspace{-2.8mm}
\footnotesize
\centering
\subfigure[\label{f5a}Medicine HyperGraph]{
    \includegraphics[width=0.31\textwidth]{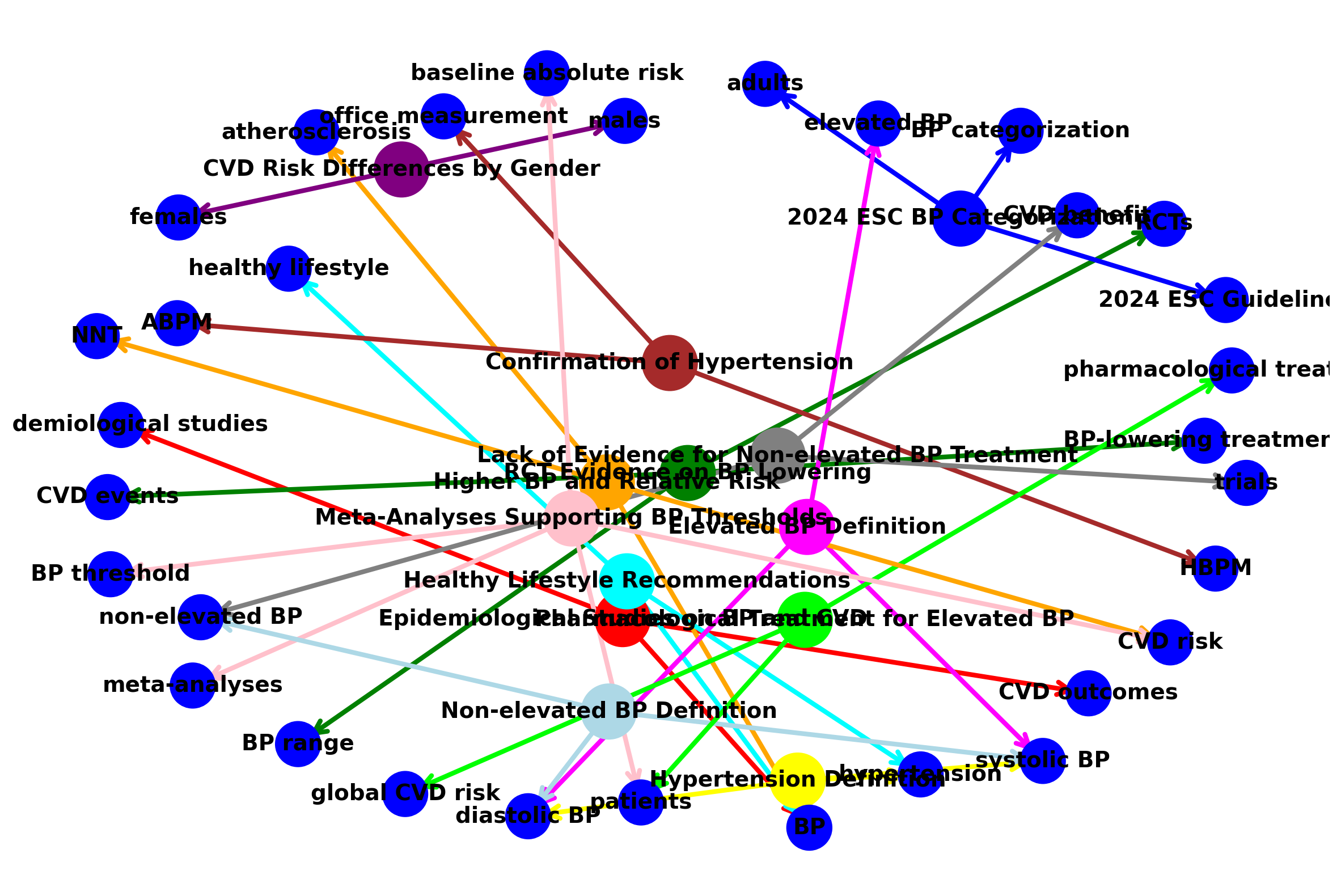}
}
\subfigure[\label{f5b}Agriculture HyperGraph]{
    \includegraphics[width=0.31\textwidth]{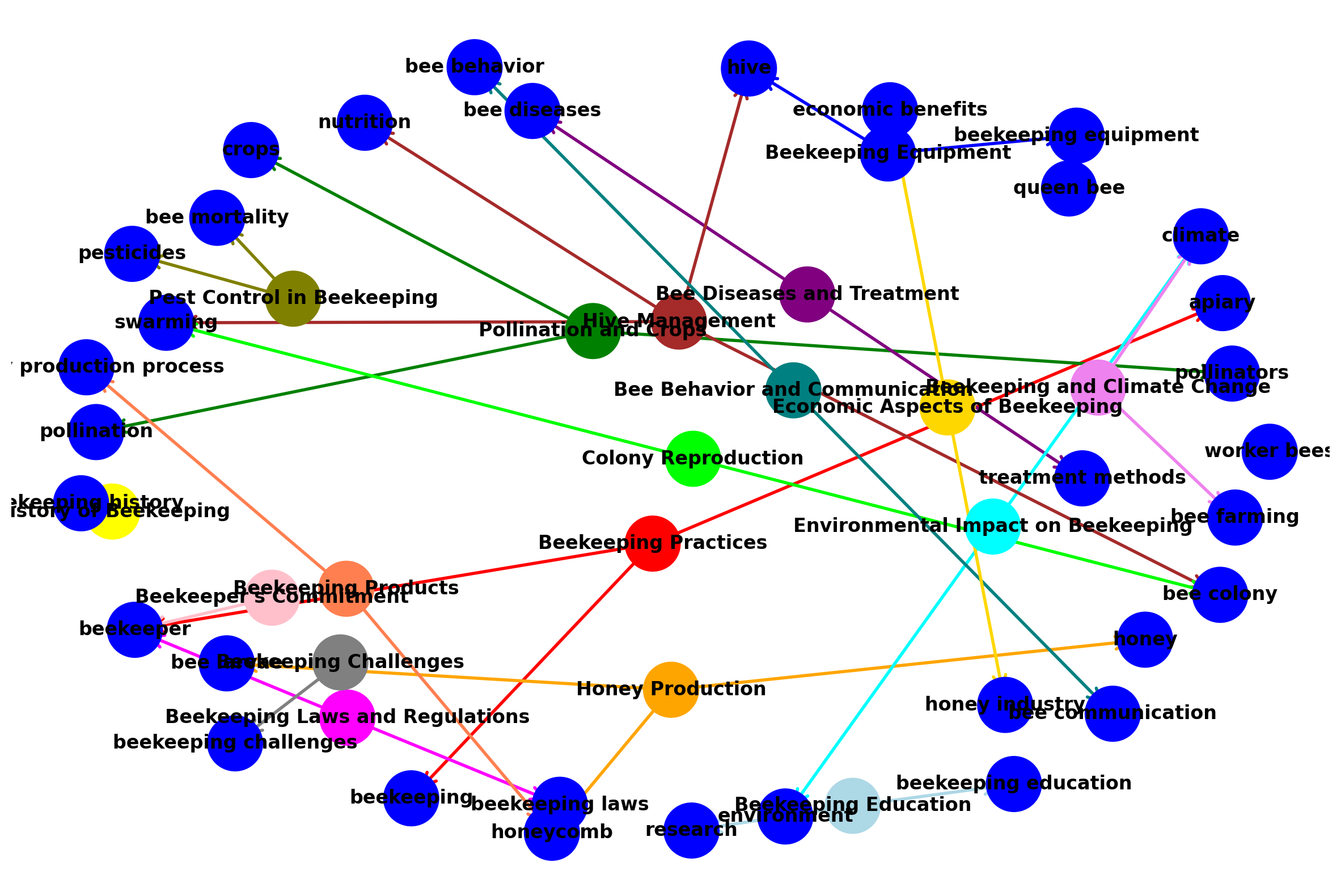}
}
\subfigure[\label{f5c}CS HyperGraph]{
    \includegraphics[width=0.31\textwidth]{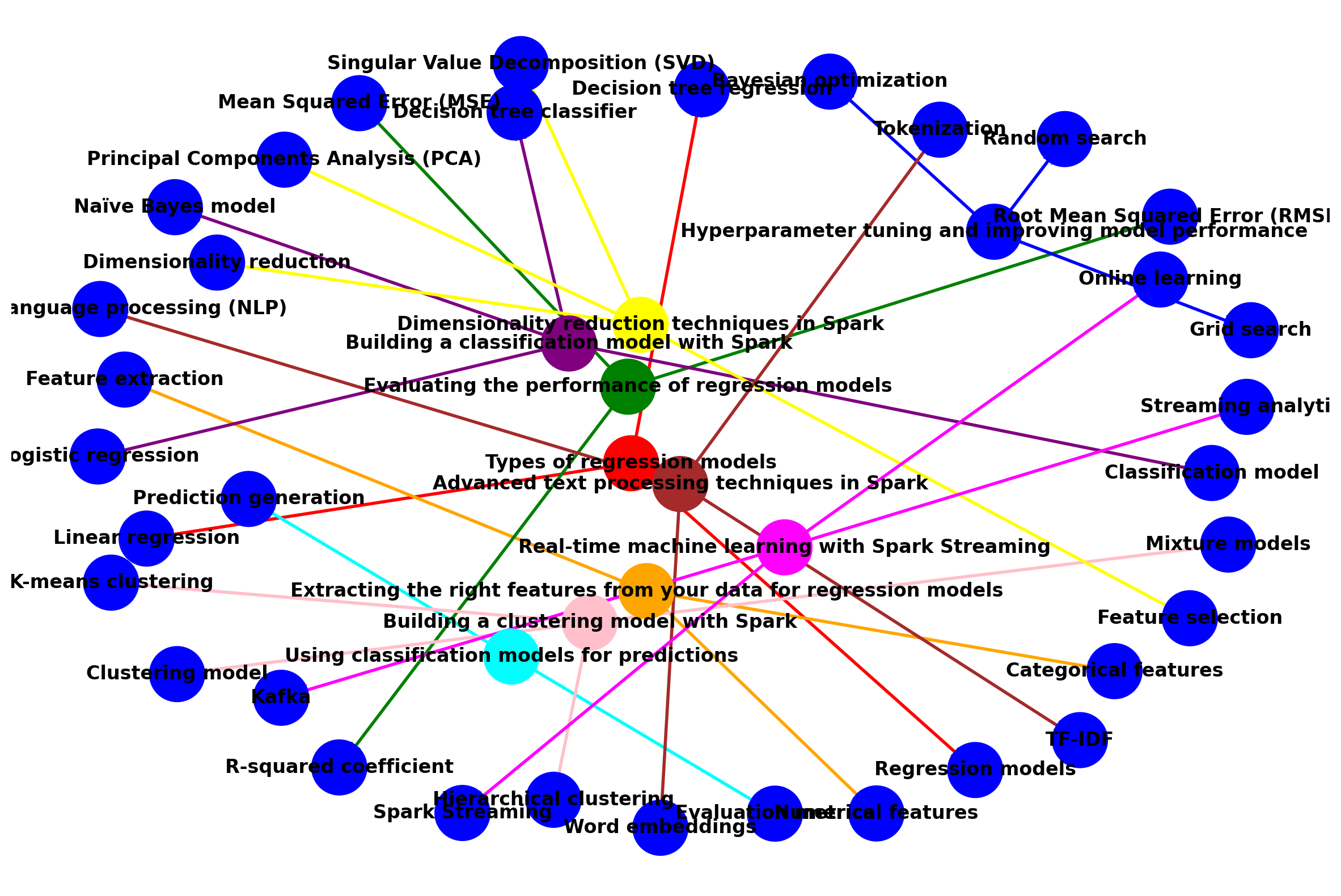}
}
\subfigure[\label{f5d}Legal HyperGraph]{
    \includegraphics[width=0.31\textwidth]{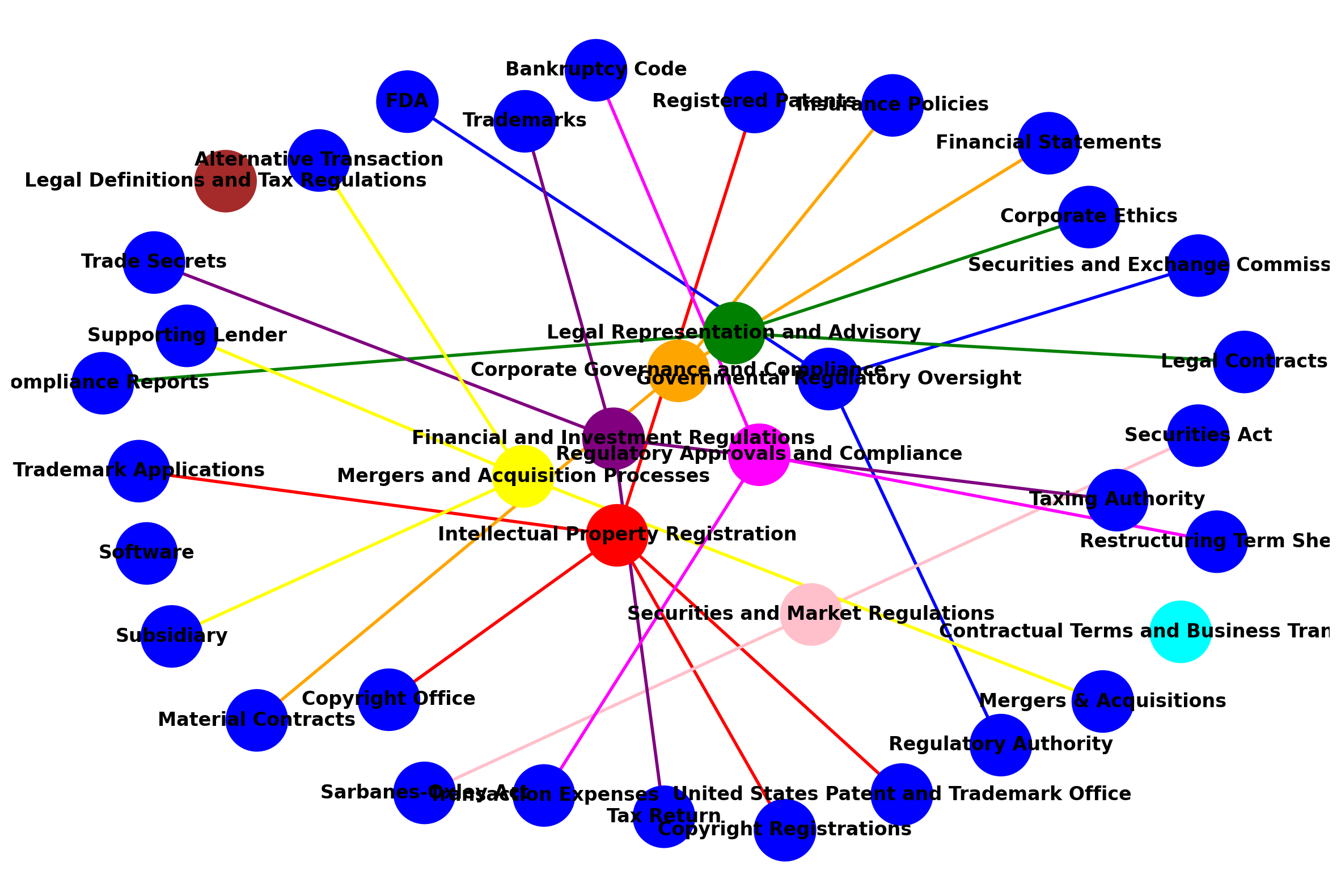}
}
\subfigure[\label{f5e}Mix HyperGraph]{
    \includegraphics[width=0.31\textwidth]{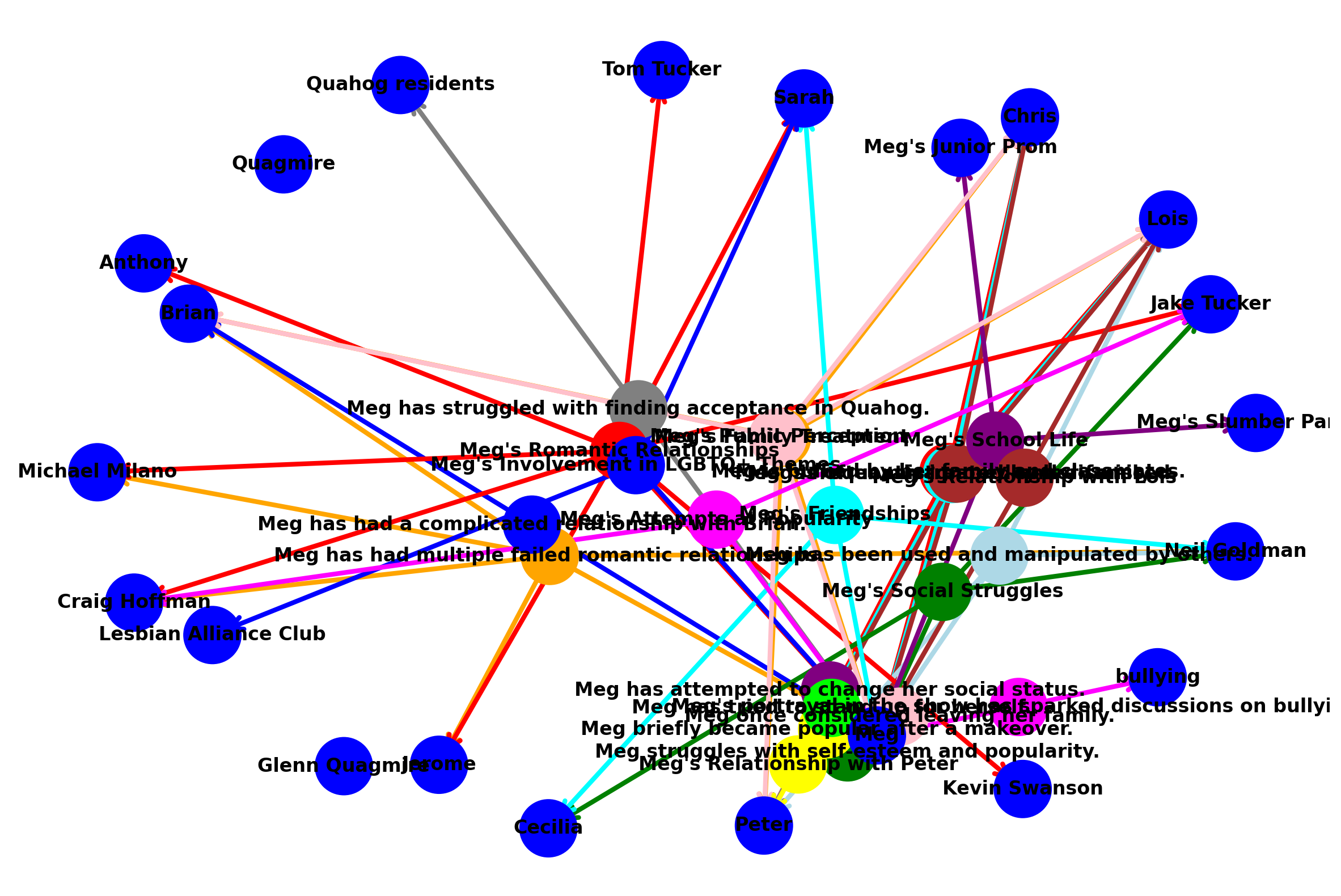}
}
\subfigure[\label{f5f}Statistics of Construction]{
    
    \begin{minipage}{0.31\textwidth}
    \vspace{-30.7mm}
        \centering
        \fontsize{5.5pt}{5.5pt}\selectfont
        \setlength{\tabcolsep}{0.45mm}{
        \begin{tabular}{lccccc}
            \toprule
             & \textbf{Med.} & \textbf{Agric.} & \textbf{CS} & \textbf{Legal} & \textbf{Mix} \\
            \midrule
            \#Knowl. Token & 179k & 382k & 795k & 940k & 122k \\
            \midrule
            \textbf{GraphRAG} & & & & & \\
            \quad \#Entity & 329 & 699 & 1449 & 1711 & 225\\
            \quad \#Community & 256 & 523 & 930 & 517 & 59\\
            \midrule
            \textbf{LightRAG} & & & & & \\
            \quad \#Entity & 3,725 & 5,032 & 8,967 & 5,354 & 2,229 \\
            \quad \#Relation & 1,304 & 3,105 & 5,632 & 6,002 & 940 \\
            \midrule
            \textbf{HyperGraphRAG} & & & & & \\
            \quad \#Entity & 7,675 & 16,805 & 19,913 & 11,098 & 6,201 \\
            \quad \#Hyperedge & 4,818 & 16,102 & 26,902 & 18,285 & 4,356 \\
            \bottomrule
        \end{tabular}}
    \end{minipage}
}
\vspace{-1mm}
\caption{(a-e) Visualizations of knowledge hypergraphs constructed in 5 domains. (f) Statistical comparison highlights HyperGraphRAG’s richer expressiveness over GraphRAG and LightRAG.}
\label{F5}
\vspace{-0.3mm}
\end{figure}

\textbf{Visualization of Knowledge Structures. }
As shown in Figure~\ref{f5a}-\ref{f5e}, unlike previous graph-based RAG methods, which only model binary relations, HyperGraphRAG connects multiple entities via hyperedges, forming a more interconnected and expressive network.

\textbf{Statistical Analysis. }
As shown in Figure~\ref{f5f}, HyperGraphRAG surpasses GraphRAG and LightRAG in all domains. For instance, in CS, it constructs 26,902 hyperedges, whereas GraphRAG has 930 communities and LightRAG 5,632 relations, showing a stronger capacity for capturing knowledge.

\subsection{Analysis of Hypergraph Retrieval Efficiency (RQ4)}
\label{5.5}
As shown in Figure~\ref{F6}, to evaluate retrieval efficiency, we conduct two experiments: (a) examining how HyperGraphRAG’s retrieval efficiency and token length scales with different top-k values and (b) comparing its F1 scores with other methods under varying retrieval length limits:
\begin{figure}[h]
\vspace{-0.5mm}
\footnotesize
\centering
\subfigure[\label{f6a}Impact of Top-k on Retreval Efficency \& Token Length.]{
    \includegraphics[width=0.545\textwidth]{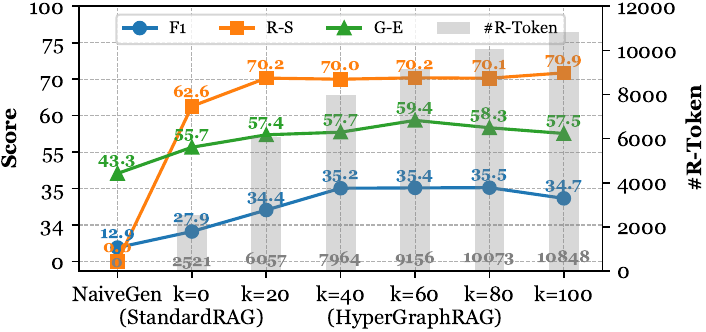}
}
\subfigure[\label{f6b}F1 Comparison under Limited Lengths.]{
    \includegraphics[width=0.4\textwidth]{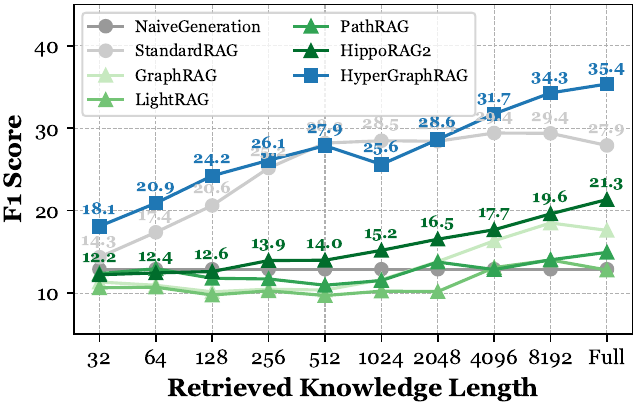}
}
\vspace{-1mm}
\caption{Experimental results in the Medicine domain analyzing hypergraph retrieval efficiency.}
\label{F6}
\vspace{-0.5mm}
\end{figure}

\textbf{Impact of Retrieved Hyperedge Quantity. }
As shown in Figure~\ref{f6a}, increasing the top-k hyperedges improves F1, R-S, and G-E, along with the rise in token count. Performance saturates around k = 60, indicating that HyperGraphRAG achieves strong retrieval quality with limited input.

\textbf{Performance under Constrained Retrieval Length. }
As illustrated in Figure~\ref{f6b}, HyperGraphRAG outperforms all binary graph-based RAG methods even under retrieval length limits, demonstrating the efficiency of n-ary representations and highlighting the semantic loss inherent in binary structures.

\subsection{Analysis of Hypergraph-Guided Generation Quality (RQ5)}
\label{5.6}
As shown in Figure~\ref{F7}, we evaluate the quality of the generation in seven dimensions:

\begin{wrapfigure}{r}{0.5\textwidth}
\vspace{-4mm}
\centering
\includegraphics[width=0.48\textwidth]{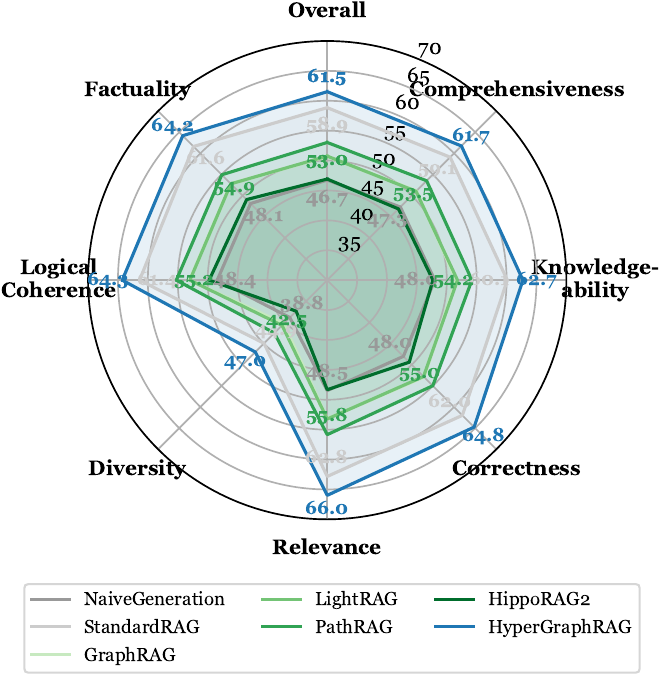}
\caption{Generation Equality Evaluations.}
\label{F7}
\vspace{-9mm}
\end{wrapfigure}

\textbf{Best Overall Generation Quality. }
HyperGraphRAG achieves the highest Overall score (61.5), significantly outperforming all baseline methods, indicating the comprehensive advantage in hypergraph-guided generation.

\textbf{Lead on Key Dimensions. }
HyperGraphRAG achieves notable improvements in Correctness (64.8), Relevance (66.0), and Factuality (64.2), outperforming both standard RAG and binary graph-based methods. These gains indicate its strong capacity to produce accurate, context-aware, and knowledge-grounded responses.

\textbf{Balanced Performance. }
Although the Diversity score (47.0) is relatively lower than other dimensions, HyperGraphRAG still exceeds all baselines, indicating that it maintains a balanced dimension-wise performance, effectively combining content richness with structural consistency for stable and high-quality generation.

\subsection{Analysis of Time and Cost in Construction and Generation Phases (RQ6)}
As shown in Table~\ref{T2}, to evaluate the efficiency and cost of HyperGraphRAG, we compare different methods in terms of knowledge construction and generation. We assess time consumption per 1k tokens (TP1kT), cost per 1k tokens (CP1kT), time per query (TPQ), and cost per 1k query (CP1kQ).

\begin{wraptable}{r}{0.5\textwidth}
\vspace{-4mm}
\centering
    \caption{\label{T2}Time \& Cost Comparisons.}
    \vspace{3mm}
    \centering
    \fontsize{8pt}{8pt}\selectfont
    \setlength{\tabcolsep}{1.4mm}{
    \begin{tabular}{lcc|cc}
        \toprule
        \multirow{2.5}{*}{\textbf{Method}} & \multicolumn{2}{c}{\textbf{Construction}} & \multicolumn{2}{c}{\textbf{Generation}} \\
        \cmidrule(lr){2-3} \cmidrule(lr){4-5}
        & \textbf{TP1kT} & \textbf{CP1kT} & \textbf{TPQ} & \textbf{CP1kQ} \\
        \midrule
        \rowcolor{gray!10} NaiveGeneration & 0 s & 0 \$ & 0.131 s & 0.059 \$ \\
        \rowcolor{gray!10} StandardRAG & 0 s &  0 \$ & 0.147 s & 1.016 \$ \\
        GraphRAG & 9.272 s & 0.0058 \$ & 0.221 s & 1.836 \$ \\
        LightRAG & 5.168 s & 0.0081 \$ & 0.359 s & 3.359 \$ \\
        PathRAG & 5.168 s & 0.0081 \$ & 0.436 s & 3.496 \$ \\
        HippoRAG2 & 2.758 s & 0.0056 \$ & 0.240 s & 3.438 \$ \\
        \rowcolor{blue!10} HyperGraphRAG & 3.084 s & 0.0063 \$ & 0.256 s & 3.184 \$ \\
        \bottomrule
    \end{tabular}}
\end{wraptable}

\textbf{Time \& Cost in Construction Phase. }
HyperGraphRAG demonstrates efficient knowledge construction with a time cost of 3.084 seconds per 1k tokens (TP1kT) and a monetary cost of \$0.0063 per 1k tokens (CP1kT). This places it between the faster HippoRAG2 (2.758s, \$0.0056) and slower GraphRAG (9.272s, \$0.0058). While its cost is slightly higher than GraphRAG, HyperGraphRAG achieves a better balance between speed, expressiveness, and structure, offering a more compact yet richer representation of n-ary relational knowledge.

\textbf{Time \& Cost in Generation Phase. }
During the generation phase, HyperGraphRAG requires 0.256 seconds per query (TPQ) and incurs a cost of \$3.184 per 1k queries (CP1kQ). This is moderately higher than StandardRAG (0.147s, \$1.016) but significantly lower than PathRAG (0.436s, \$3.496) and LightRAG (0.359s, \$3.359). Compared to GraphRAG (0.221s, \$1.836), HyperGraphRAG slightly increases time and cost but compensates with better retrieval quality and generation outcomes. The results suggest that HyperGraphRAG achieves a favorable trade-off between generation efficiency and output quality, suitable for real-world knowledge-intensive applications.

\section{Conclusion}
In this work, we present HyperGraphRAG, a retrieval-augmented generation framework that models knowledge as hypergraphs to capture n-ary relational structures. By introducing novel methods for knowledge hypergraph construction, retrieval, and generation, HyperGraphRAG addresses limitations of binary graph-based RAG methods. Experimental results across diverse domains demonstrate consistent improvements in answer accuracy, retrieval relevance, and generation quality, confirming the effectiveness and generalizability of hypergraph-guided retrieval and generation.

\section*{Acknowledgments}
This work is supported by the National Natural Science Foundation of China (Grant No. 62473271, Grant No. 62176026, and Grant No. 62406036) and the Engineering Research Center of Information Networks, Ministry of Education, China.

\bibliography{custom}{}
\bibliographystyle{plain}


\newpage
\appendix
\section*{Appendix} 
\section{Prompts Used in HyperGraphRAG}

\subsection{N-ary Relation Extraction Prompt}
\label{AppendixA1}
\begin{figure}[h]
\vspace{-1mm}
\centering
\includegraphics[width=0.98\linewidth]{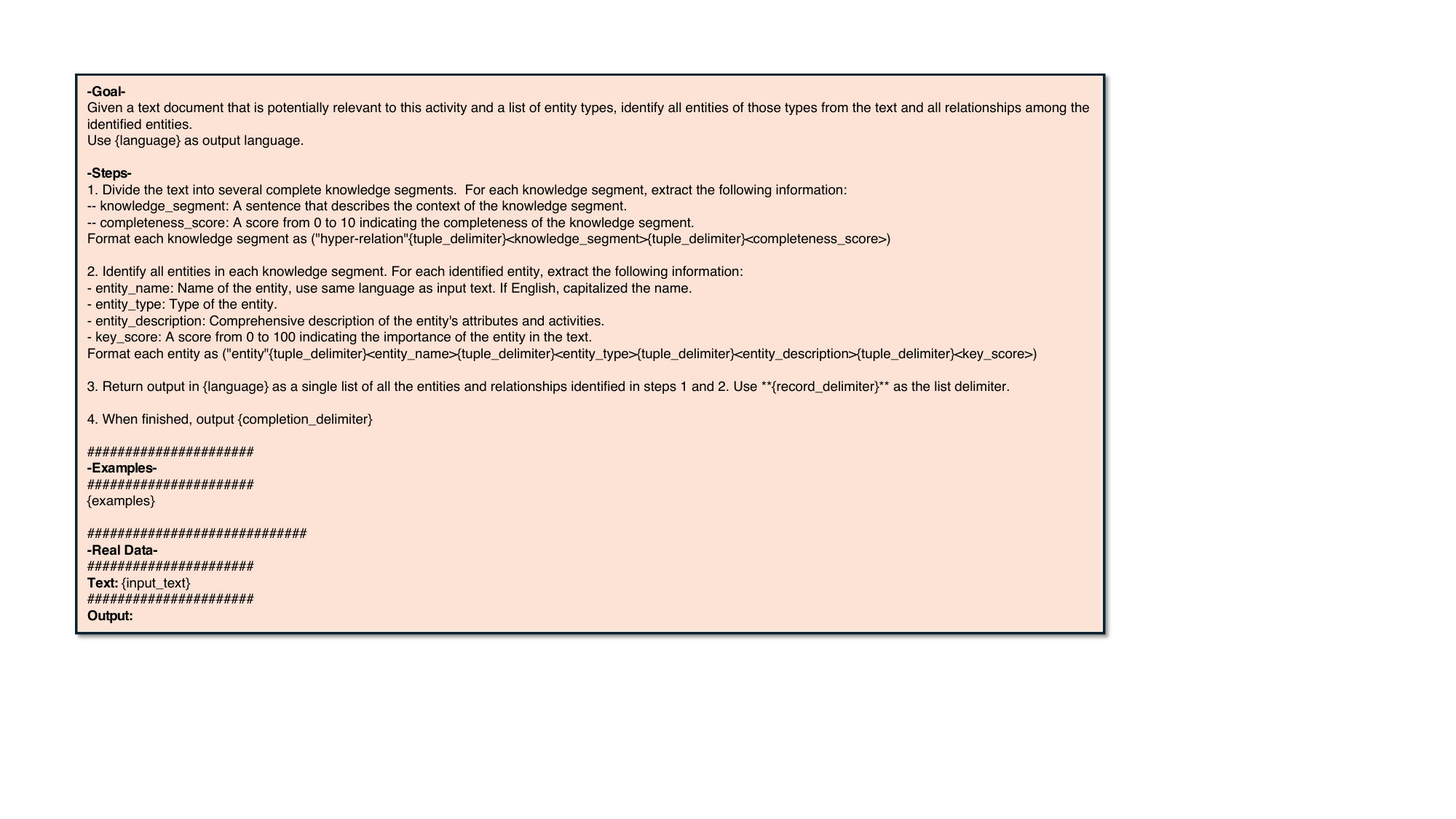}
\vspace{-1mm}
\caption{\label{prompt1}
Prompt for n-ary relation extraction $p_{\text{ext}}$ in Equation~\ref{E4}.}
\vspace{-5mm}
\end{figure}

As shown in Figure~\ref{prompt1}, this prompt is designed for extracting structured n-ary relational facts from raw text. It guides LLM to segment the input into coherent knowledge fragments, assign a completeness score to each, and identify entities with their names, types, descriptions, and importance scores.

\subsection{Entity Extraction Prompt}
\label{AppendixA2}
\begin{figure}[h]
\vspace{-1mm}
\centering
\includegraphics[width=0.98\linewidth]{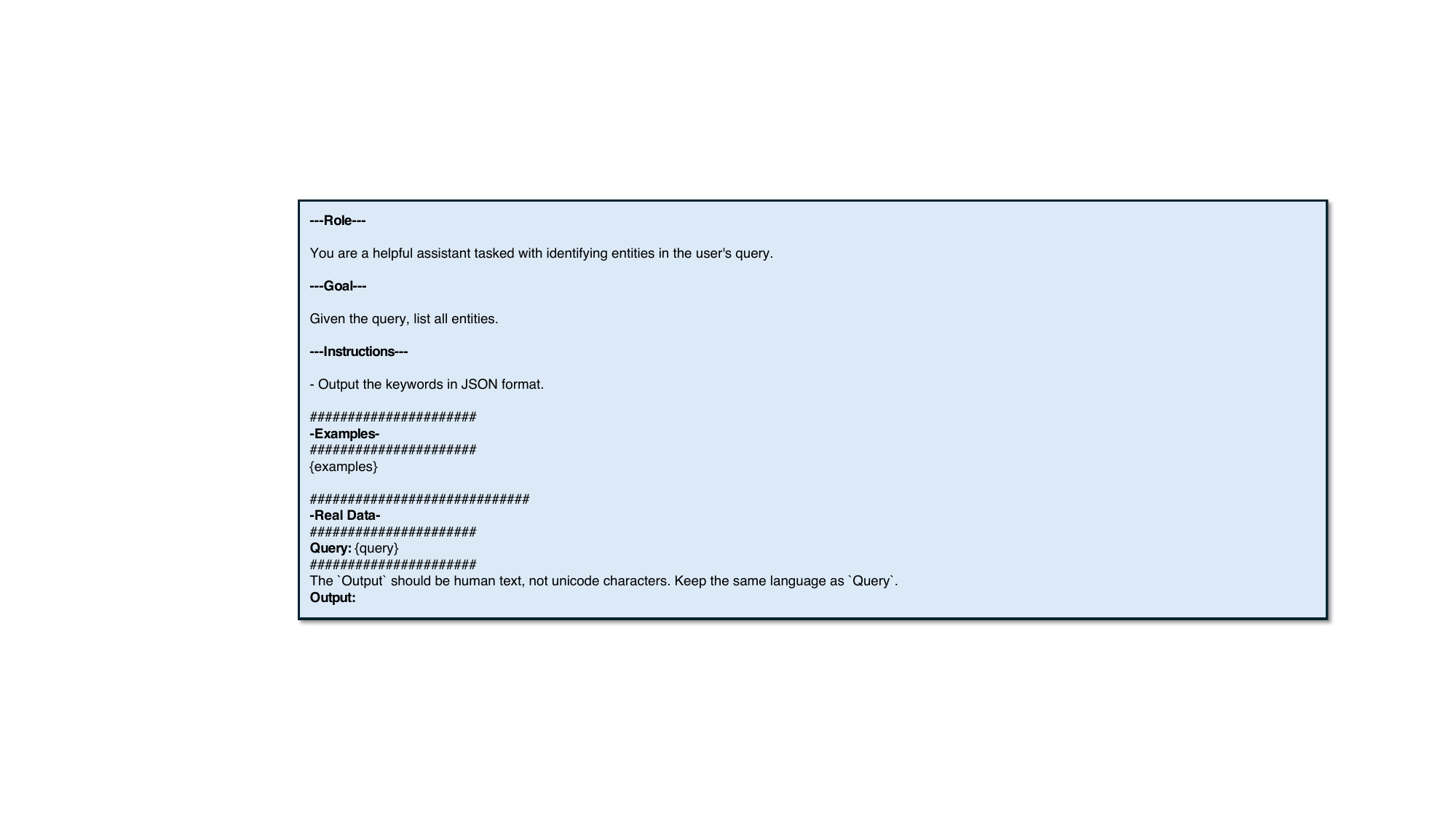}
\vspace{-1mm}
\caption{\label{prompt2}
Prompt for entity extraction $p_{\text{q\_ext}}$ in Equation~\ref{E7}.}
\vspace{-5mm}
\end{figure}

As shown in Figure~\ref{prompt2}, this prompt is used to extract key entities from a user query. LLM is instructed to return all identified entities in JSON format, ensuring the output is concise, human-readable, and aligned with the language of the input query. This facilitates entity-level retrieval in the hypergraph.

\subsection{Retrieval-Augmented Generation Prompt}
\label{AppendixA3}
To ensure a fair comparison across RAG baselines, we adopt a unified Chain-of-Thought (CoT)-based generation prompt $p_{\text{gen}}$ in Equation~\ref{E12} for all methods. We present this prompt together with the designed evaluation approach in Appendix~\ref{AppendixE}.

\section{Proof}
\label{proofs}
\subsection{Proof of Proposition 1}
\label{proof1}
\textbf{Proposition 1.}
\textit{Hypergraph-structured knowledge representation is more comprehensive than binary.}\vspace{-2mm}
\begin{proof} 
Given a universe of entities $V$, an $n$-ary fact with $n \ge 3$ is denoted as $F = \{v_1, \ldots, v_n\} \subseteq V$.
For hypergraph representation, we represent it with a single hyperedge:
\begin{equation}
e_H = F,\quad G_H = (V, E_H),\quad e_H \in E_H,
\end{equation}
so the representation function $\phi_H : F \mapsto e_H$ is naturally injective.
For binary graph representation, we connect every pair of entities that co-occur in a fact. For any collection of facts $\mathcal S \subseteq \mathcal P(V)$, define the representation function:
\begin{equation}
\phi_B(\mathcal S) = (V_B, E_B), \quad
V_B = \bigcup_{F \in \mathcal S} F,\quad
E_B = \left\{(u,w) \mid u \ne w,\; \exists F \in \mathcal S:\; \{u,w\} \subseteq F \right\},
\end{equation}
where $E_B$ consists of the binary edges activated by $\mathcal S$ within the complete graph $K_{|V|}$. Each $E_B$ is a subset of some clique.

Let the random variable $X$ range over all possible fact sets $\mathcal S$, with Shannon entropy:
\begin{equation}
H(X) = -\sum_{\mathcal S} p(\mathcal S)\log_2 p(\mathcal S),
\end{equation}
measuring the total information to be represented.
For hypergraph representation, since $\phi_H$ is injective and each fact can be uniquely recovered,
\begin{equation}
H(X \mid \phi_H(X)) = 0.
\end{equation}
For binary representation, consider any three distinct entities $a, b, c \in V$, and define
\begin{equation}
\mathcal S_1 = \{\{a,b,c\}\}, \quad
\mathcal S_2 = \bigl\{\{a,b\},\; \{a,c\},\; \{b,c\}\bigr\}.
\end{equation}
Clearly, $\mathcal S_1 \ne \mathcal S_2$, but
\begin{equation}
\phi_B(\mathcal S_1) = \phi_B(\mathcal S_2) = \bigl(\{a,b,c\},\; \{(a,b),(a,c),(b,c)\}\bigr)=g,
\end{equation}
since both activate the same set of binary edges. Thus,
$$
\left|\phi_B^{-1}(\phi_B(\mathcal S_1))\right| \ge 2,
\quad\Rightarrow\quad
0<\frac{p(\mathcal S_i)}{p(g)}<1,
$$
\begin{equation}
\quad\Rightarrow\quad
H(X \mid \phi_B(X)=g) = - \sum_{\mathcal S_i \in \phi_B^{-1}(g)} \frac{p(\mathcal S_i)}{p(g)} \log_2 \frac{p(\mathcal S_i)}{p(g)} > 0,
\end{equation}
then, we can get
\begin{equation}
H(X \mid \phi_B(X)) =\sum_y p(y) H(X \mid \phi_B(X) = y) \geq p(g)H(X \mid \phi_B(X)=g) > 0,
\end{equation}
where information is inevitably lost in binary representation.

More generally, as long as there exists at least one $n$-ary fact ($n \ge 3$) in the knowledge base, we can always construct a pair of distinct fact sets that activate the same binary edges through a merge-split transformation. Hence,
\begin{equation}
H(X \mid \phi_B(X)) > 0,
\quad
I(X ; \phi_B(X)) = H(X) - H(X \mid \phi_B(X)) < H(X),
\end{equation}
which proves that binary representation is lossy.
In contrast, hypergraph representation satisfies $H(X \mid \phi_H(X)) = 0$, so the mutual information reaches its upper bound $H(X)$ and all information is preserved.
In the special case where no $n$-ary facts with $n \ge 3$ exist, i.e., all facts are binary, then
\begin{equation}
\left|\phi_B^{-1}(g_B)\right| = 1, \quad
H(X \mid \phi_B(X)) = 0,
\end{equation}
so binary representation becomes injective and equivalent to hypergraph, with no information loss.

In conclusion, as long as the knowledge base contains at least one fact of arity three or higher, hypergraph-structured representation preserves more information with lossless representation, whereas binary representation inevitably loses information. Therefore, hypergraph representation is more comprehensive than binary in the information-theoretic sense.
\end{proof}
\subsection{Proof of Proposition 2}
\label{proof2}
\textbf{Proposition 2.}
\textit{A bipartite graph can losslessly preserve and query a knowledge hypergraph.}\vspace{-2mm}
\begin{proof} 
Let the knowledge hypergraph be denoted as $G_H = (V, E_H), E_H \subseteq \{\, e_H \subseteq V \mid |e_H| \ge 2 \}$.
Each hyperedge is abstracted as a new node, and combined with the set of entity nodes to form a new vertex set $V_B = V \cup E_H$, with edges defined as $E_B = \{\, (e_H, v) \mid e_H \in E_H,\; v \in e_H \}$, resulting in the incidence bipartite graph $\Phi(G_H) = G_B = (V_B, E_B)$.

Ordering the vertices such that entities come first and hyperedges second, $G_H$ can be represented by the binary incidence matrix  
\begin{equation}
M \in \{0,1\}^{|V| \times |E_H|}, \quad M_{v,e_H} = 1 \iff v \in e_H,
\end{equation}
and the adjacency matrix of $G_B$ becomes  
\begin{equation}
\label{EP1}
A_{G_B} =
\begin{pmatrix}
0 & M \\
M^{\!\top} & 0
\end{pmatrix}
\end{equation}
where $M$ uniquely determines $A_{G_B}$, and conversely, $M$ can be recovered from the top-right block of $A_{G_B}$. Therefore, there exists an inverse mapping: 
\begin{equation}
\Phi^{-1}: G_B \to G_H, \quad
\Phi^{-1}(V_B, E_B) = \left(V, \{\, N_{G_B}(e_H) \mid e_H \in E_H \} \right),
\end{equation}
where  
\begin{equation}
N_{G_B}(e_H) = \{ v \in V \mid (e_H, v) \in E_B \}.
\end{equation}
Clearly,  
\begin{equation}
\Phi^{-1} \circ \Phi = \mathrm{id}_{G_H}, \quad 
\Phi \circ \Phi^{-1} = \mathrm{id}_{G_B},
\end{equation}
which means that $\Phi$ is a bijection and the encoding is lossless.

The query equivalence can also be derived directly via matrix operations and path counting: the set of hyperedges containing an entity $v$ corresponds to the support of the $v$-th row of $M$, and in the bipartite graph this is equivalent to the neighborhood $N_{G_B}(v)$, given by the right block of $\mathbf{e}_v^{\!\top} A_{G_B} = (0,\; \mathbf{e}_v^{\!\top} M)$. Likewise, the entity set of a hyperedge $e_H$ is the support of the $e_H$-th column of $M$, which matches the left block of $\mathbf{f}_{e_H}^{\!\top} A_{G_B}$. To determine whether two entities $u, v$ co-occur in some hyperedge, it suffices to check whether  
\begin{equation}
(MM^{\!\top})_{uv} = (A_{G_B}^2)_{uv} \neq 0,
\end{equation}
since $(A_{G_B}^2)_{uv}$ counts all 2-step paths from $u$ through a hyperedge node to $v$. For a given subset of entities $S \subseteq V$, hyperedges that contain all of them can be found by summing the corresponding rows $\sum_{v \in S} M_{v,*}$ and selecting columns where the sum equals $|S|$; in the bipartite graph, this corresponds to the intersection  
\begin{equation}
\bigcap_{v \in S} N_{G_B}(v).
\end{equation}
All operations run in time $O(|E_B|)$, which matches the complexity of equivalent queries over $G_H$.

In conclusion, the bijection $\Phi$ guarantees full structural reversibility, while adjacency and path-based reasoning preserve the semantics of all queries involving entity–hyperedge membership. Therefore, a bipartite graph can losslessly preserve and query a knowledge hypergraph.
\end{proof}

\subsection{Proof of Proposition 3}
\label{proof3}
\textbf{Proposition 3.}
\textit{Retrieving knowledge on a knowledge hypergraph improves retrieval efficiency compared to methods based on ordinary binary graphs, leading to gains in generation quality.}\vspace{-2mm}
\begin{proof} 
Let the ground-truth knowledge set required for a query $q$ be modeled as a discrete random variable $X \subseteq \mathcal{P}(V)$, with probability measure $\mu$ defined over the measurable space $(\mathcal{P}(V), \mathcal{B})$. For any $n$-ary fact $F = \{v_1, \dots, v_n\}$ with $n \ge 3$, we define two encoders:
\begin{equation}
\varphi_H\colon F \longmapsto e_H = F, \quad
\varphi_B\colon F \longmapsto \{(v_i, v_j)\mid 1 \le i < j \le n \}.
\end{equation}
Let the encoded knowledge sets be random variables $Y_H = \varphi_H(X)$ and $Y_B = \varphi_B(X)$. Since $\varphi_H$ is injective, the conditional entropy is zero:
\begin{equation}
\label{PE1}
H\left(X \mid Y_H\right) = 0,
\quad\text{and hence}\quad
I(X; Y_H) = H(X).
\end{equation}
However, when $\mu(\{|F| \ge 3\}) > 0$, the encoder $\varphi_B$ becomes non-injective. There exist $x_1 \ne x_2$ such that $Y_B(x_1) = Y_B(x_2)$, leading to:
\begin{equation}
H\left(X \mid Y_B\right)
= \mathbb{E}_{Y_B} \left[ -\sum_{x \in \varphi_B^{-1}(Y_B)} \mu(x \mid Y_B)\log_2 \mu(x \mid Y_B) \right] > 0,
\end{equation}
\begin{equation}
\label{PE2}
I(X; Y_B) = H(X) - H(X \mid Y_B) < H(X).
\end{equation}
To study encoding efficiency, consider encoding $Y_\star$ ($\star \in \{H, B\}$) using an optimal prefix code. Let the expected code length be $\mathcal{L}_\star = \mathbb{E}[\ell(Y_\star)]$. According to Shannon's source coding theorem:
\begin{equation}
\mathcal{L}_\star \in [H(Y_\star), H(Y_\star) + 1).
\end{equation}
Define the information efficiency density (information per bit) as:
\begin{equation}
\eta_\star = \frac{I(X; Y_\star)}{\mathcal{L}_\star}.
\end{equation}
This metric quantifies the amount of effective information transmitted per bit. Since $I(X; Y_H) = H(X)$ while $I(X; Y_B) < H(X)$, and $H(Y_B) \ge H(Y_H)$ (as the pairwise representation introduces a larger outcome space), we have:
\begin{equation}
\eta_H - \eta_B
= \frac{H(X)}{\mathcal{L}_H} - \frac{H(X) - \delta}{\mathcal{L}_B},
\quad \delta > 0, \quad
\mathcal{L}_B - \mathcal{L}_H \ge 0,
\end{equation}
which is strictly positive when $\delta > 0$. This shows that the hypergraph representation transmits more effective information per bit.
Let the maximum retrievable context budget for a language model be $L$, and define the coverage function:
\begin{equation}
\mathcal{C}_\star(L) = \Pr\bigl(I(X; Y_\star) \le L\bigr)
= \mu\left(\{x \mid \eta_\star \cdot \ell(Y_\star(x)) \le L\}\right),
\end{equation}
$\mathcal{C}_\star(L)$ is a non-decreasing function of $L$ and is differentiable almost everywhere. Given $\eta_H > \eta_B$, the chain rule yields:
\begin{equation}
\label{PE4}
\frac{d}{dL} \mathcal{C}_H(L)
= \int_{\ell(Y_H) = L/\eta_H}
\frac{\partial \mu}{\partial \ell}
\cdot \frac{d\ell}{dL} \,d\sigma
\ge
\int_{\ell(Y_B) = L/\eta_B}
\frac{\partial \mu}{\partial \ell}
\cdot \frac{d\ell}{dL} \,d\sigma
= \frac{d}{dL} \mathcal{C}_B(L),
\end{equation}
which implies $\mathcal{C}_H(L) \ge \mathcal{C}_B(L)$ with strict inequality on intervals where $\mu(\{|F| \ge 3\}) > 0$.
Let generation quality $E$ (e.g., G-E score) be a differentiable function $E = g(I(X; Y_\star), \mathcal{N}_\star)$, where $\mathcal{N}_\star$ denotes the noise introduced by irrelevant or redundant edges, and satisfies:
\begin{equation}
\frac{\partial g}{\partial I} > 0, \quad
\frac{\partial g}{\partial \mathcal{N}} < 0.
\end{equation}
Here, noise $\mathcal{N}_\star$ is defined as the set of edges retrieved under budget $L$ that are irrelevant to the ground-truth $X^\dagger$. Under the same bit budget, higher $\eta_H$ implies fewer edges per bit, and thus:
\begin{equation}
\label{PE5}
\mathbb{E}[\mathcal{N}_H] \le \mathbb{E}[\mathcal{N}_B].
\end{equation}
Treating $L$ as an independent variable, we apply the chain rule:
\begin{equation}
\frac{d}{dL}\bigl[E_H(L) - E_B(L)\bigr]
= \frac{\partial g}{\partial I}(\theta_L)
\left[\frac{d}{dL} I(X; Y_H) - \frac{d}{dL} I(X; Y_B)\right]
+
\frac{\partial g}{\partial \mathcal{N}}(\theta_L)
\left[\frac{d}{dL} \mathcal{N}_H - \frac{d}{dL} \mathcal{N}_B\right],
\end{equation}
where $\theta_L$ is an intermediate state between the two systems. From Equation~\ref{PE4} and Equation~\ref{PE5}, we know: (1) The first term is strictly positive if high-arity facts exist; (2) The second term is always non-positive, as higher information density leads to lower redundancy.
Therefore, the total derivative is strictly positive. Integrating over $[0, L]$, we obtain:
\begin{equation}
\label{PE6}
E_H(L) - E_B(L)
= \int_0^L \frac{d}{d\beta} \left[E_H(\beta) - E_B(\beta)\right]\,d\beta > 0,
\quad\text{unless } \mu(\{|F| \ge 3\}) = 0.
\end{equation}
Equation~\ref{PE6} formally proves that if there exists at least one fact with arity $n \ge 3$ in the knowledge base, then under any fixed retrieval budget $L$, the generation quality under hypergraph encoding strictly exceeds that of the binary encoding. In the degenerate case where all facts are binary, both encodings reduce to the same mapping, and the conclusion naturally becomes an equality.
\end{proof}

\section{HyperGraphRAG Algorithm Detail}

\textbf{Hypergraph Construction.}
To provide a clear overview of our system pipeline, we present the detailed procedures of HyperGraphRAG in the form of pseudocode. As shown in Algorithm~\ref{alg:build_hypergraph}, we first construct a knowledge hypergraph from raw documents via LLM-based extraction of n-ary relational facts. Each extracted fact forms a hyperedge connecting multiple entities, and the resulting hypergraph is stored in a bipartite structure for efficient indexing and retrieval. We further compute dense embeddings for all entities and hyperedges to support semantic retrieval.

\begin{algorithm}[h!t]
\small
\caption{Hypergraph Construction}
\label{alg:build_hypergraph}
\begin{algorithmic}[1]
\REQUIRE Document collection $\mathcal{D}$
\ENSURE Knowledge hypergraph $\mathcal{G}_H = (V, E_H)$

\STATE Initialize entity set $V \leftarrow \emptyset$, hyperedge set $E_H \leftarrow \emptyset$
\FOR{each document $d \in \mathcal{D}$}
    \STATE Extract n-ary facts: $\mathcal{F}_d = \{(e_i, V_{e_i})\}_{i=1}^{k} \sim \pi(d)$
    \STATE $V \leftarrow V \cup \bigcup_{i=1}^k V_{e_i}$
    \STATE $E_H \leftarrow E_H \cup \{e_i\}_{i=1}^{k}$
\ENDFOR
\STATE Store $(V, E_H)$ as bipartite graph $\mathcal{G}_B = \Phi(\mathcal{G}_H)$
\STATE Compute embeddings: $E_V = \{f(v) \mid v \in V\}$, $E_{E_H} = \{f(e) \mid e \in E_H\}$
\RETURN $\mathcal{G}_H = (V, E_H)$
\end{algorithmic}
\end{algorithm}

\textit{Complexity Analysis.} 
Given a corpus of $D$ documents, assume each document contains at most $r$ relational facts, and each fact involves up to $n$ entities. The LLM-based extraction step has complexity $\mathcal{O}(D)$ under the assumption of constant-time per document prompt. Constructing the hypergraph involves inserting up to $\mathcal{O}(D \cdot r)$ hyperedges and $\mathcal{O}(D \cdot r \cdot n)$ entities (with deduplication), resulting in a total construction time of $\mathcal{O}(D \cdot r \cdot n)$. Embedding all nodes and hyperedges requires $\mathcal{O}(|V| + |E_H|)$ calls to the encoder, typically parallelizable.
\vspace{2mm}

\textbf{Hypergraph Retrieval and Generation.} 
Once the hypergraph is constructed, the generation process begins with a query input, as detailed in Algorithm~\ref{alg:retrieval_generation}. We first extract relevant entities from the query and perform top-$k$ similarity search to retrieve both entity and hyperedge candidates. We then perform bidirectional neighborhood expansion over the hypergraph to assemble a knowledge set, which may optionally be combined with chunk-level retrieval. Finally, we format the retrieved knowledge into a prompt and generate an answer using a large language model. This modular pipeline ensures efficient, expressive, and accurate generation grounded in structured knowledge.

\begin{algorithm}[h!t]
\small
\caption{Hypergraph Retrieval and Generation}
\label{alg:retrieval_generation}
\begin{algorithmic}[1]
\REQUIRE Query $q$, knowledge hypergraph $\mathcal{G}_H = (V, E_H)$
\ENSURE Final answer $y^*$

\STATE Extract query entities: $V_q \sim \pi(q)$
\STATE Retrieve top-$k$ entities: $V_r \leftarrow \textsc{TopKSim}(V_q, E_V)$
\STATE Retrieve top-$k$ hyperedges: $E_r \leftarrow \textsc{TopKSim}(q, E_{E_H})$
\STATE Expand neighbors: $F^*_V = \bigcup_{v \in V_r} \text{Nbr}(v), \quad F^*_E = \bigcup_{e \in E_r} \text{Nbr}(e)$
\STATE Assemble retrieved knowledge: $K_H = F^*_V \cup F^*_E$
\STATE Retrieve additional chunks (optional): $K_{\text{chunk}} = \textsc{RetrieveChunks}(q)$
\STATE Combine all knowledge: $K^* = K_H \cup K_{\text{chunk}}$
\STATE Generate answer: $y^* \sim \pi(q, K^*)$
\RETURN $y^*$
\end{algorithmic}
\end{algorithm}

\textit{Complexity Analysis.} 
Given a query $q$, entity and hyperedge retrieval involves computing top-$k$ similarity against all entity and hyperedge embeddings. With $|V|$ entities and $|E_H|$ hyperedges, this results in $\mathcal{O}(|V| + |E_H|)$ embedding comparisons. The neighborhood expansion step is bounded by the degree of retrieved nodes, i.e., $\mathcal{O}(k \cdot d)$ where $d$ is average node degree. Finally, generation is treated as a black-box LLM inference, typically $\mathcal{O}(L)$ where $L$ is the prompt length.

In summary, HyperGraphRAG achieves efficient inference with precomputed indices, and its overall retrieval-generation time is dominated by vector similarity lookup and prompt generation, both of which scale linearly with hypergraph size and are highly parallelizable in practice.

\section{Dataset Construction}
\label{AppendixA}
\subsection{Knowledge Domains}
The dataset used for HyperGraphRAG evaluation covers five domains, with data sourced as follows:

\textbf{Medicine:} Derived from the latest international hypertension guidelines~\citep{Hypertension}, covering medical diagnosis, treatment plans, and clinical indicators.
\textbf{Agriculture:} Extracted from the UltraDomain dataset~\citep{UltraDomain}, including knowledge on agricultural production, crop management, and pest control.
\textbf{Computer Science (CS):} Sourced from the UltraDomain dataset, encompassing computer architecture, algorithms, and machine learning.
\textbf{Legal:} Based on the UltraDomain dataset, covering legal provisions, judicial precedents, and regulatory interpretations.
\textbf{Mix:} A combination of multiple domains to assess the model’s generalization ability across interdisciplinary tasks.

\subsection{Question Sampling Strategies}
To construct a fair and comprehensive evaluation benchmark, we design a uniform sampling strategy for both binary and n-ary sources. Specifically, for each domain, we sample a total of \textbf{512 questions}, consisting of:

\textbf{Binary Source (256 samples):} 
128 facts are selected via 1-hop traversal, 64 facts via 2-hop traversal, 64 facts via 3-hop traversal. These facts are composed of binary relations (i.e., pairwise entity connections) and are used to build the binary knowledge source.

\textbf{N-ary Source (256 samples):} 
128 facts are sampled via 1-hop traversal, 64 facts via 2-hop traversal, 64 facts via 3-hop traversal. These facts involve multi-entity ($n \geq 3$) relational structures and are used to construct the n-ary knowledge source.

For each sampled fact, we prompt GPT to generate a corresponding question and its golden answer. All generated question-answer pairs are manually verified to ensure factual accuracy, relevance, and diversity. This process is repeated independently for every domain to ensure consistent scale and structure across evaluation sets. All datasets undergo manual review to ensure the accuracy of annotated answers and the fairness of model evaluation.

\section{Evaluation Details}
\label{AppendixE}
\begin{figure}[h]
\centering
\includegraphics[width=0.98\linewidth]{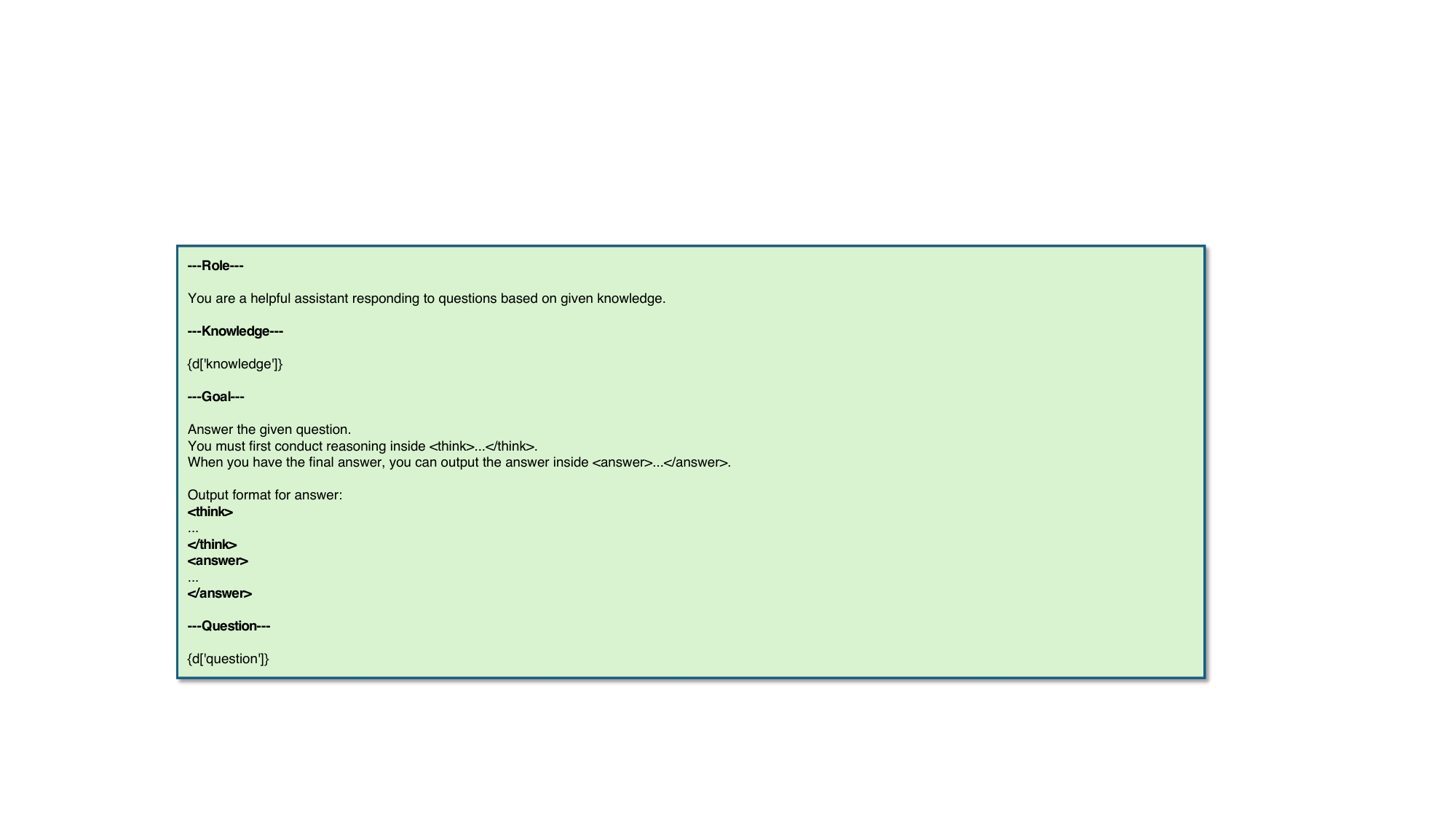}
\caption{\label{prompt3}
The unified prompt for generation $p_{\text{gen}}$ in Equation~\ref{E12}.}
\end{figure}

\textbf{Unified Generation Prompt.} To ensure a fair comparison across all baselines, we adopt a unified generation prompt for all methods, as shown in Figure~\ref{prompt3}. Specifically, we insert the knowledge retrieved by each method into a fixed prompt template that guides the model to first perform reasoning within a \texttt{<think>} block and then provide the final answer within an \texttt{<answer>} block, preserving benefits of zero-shot CoT reasoning while maintaining consistency across different retrieval strategies.

We evaluate model performance using three complementary metrics that assess different aspects of retrieval-augmented generation: factual alignment, retrieval quality, and generation fluency.

\textbf{(i) F1 Score.} 
Following FlashRAG~\citep{FlashRAG}, we compute the word-level F1 score between each generated answer and its ground-truth reference, and then average over all questions. This metric captures reflects factual alignment with the expected answer.
\begin{equation}
\text{F1} = \frac{1}{N} \sum_{i=1}^{N} \frac{2 \cdot P_i \cdot R_i}{P_i + R_i}, \quad
P_i = \frac{|\text{Pred}_i \cap \text{GT}_i|}{|\text{Pred}_i|}, \quad
R_i = \frac{|\text{Pred}_i \cap \text{GT}_i|}{|\text{GT}_i|}
\end{equation}
where $\text{Pred}_i$ and $\text{GT}_i$ denote the set of words in the predicted and ground-truth answers for the $i$-th question, and $N$ is the total number of evaluated questions.

\textbf{(ii) Retrieval Similarity (R-S).} 
Inspired by RAGAS~\citep{RAGAS}, R-S quantifies the semantic similarity between the retrieved knowledge and the ground-truth knowledge used to construct the question. For each question, we concatenate all retrieved knowledge into a single string $k_{\text{retr}}$ and all golden knowledge into $k_{\text{gold}}$, then compute the cosine similarity between their embeddings. The final R-S score is the average similarity across the dataset:
\begin{equation}
\text{R-S} = \frac{1}{N} \sum_{i=1}^{N} \cos\left( f(k^{(i)}_{\text{retr}}),\; f(k^{(i)}_{\text{gold}}) \right)
\end{equation}
where $f(\cdot)$ is the embedding function (e.g., SimCSE), and $N$ is the total number of questions.

\textbf{(iii) Generation Evaluation (G-E).}
Adapted from HelloBench~\citep{HelloBench}, G-E uses GPT-4o-mini as an LLM judge to evaluate generation quality along seven dimensions: \textit{Correctness}, \textit{Relevance}, \textit{Factuality}, \textit{Comprehensiveness}, \textit{Knowledgeability}, \textit{Logical Coherence}, and \textit{Diversity}. For each question, we compute the average of the seven dimension scores, then combine it with the question’s F1 score by taking their mean. The final G-E score is obtained by averaging this combined score:
\begin{equation}
\text{G-E} = \frac{1}{N} \sum_{i=1}^{N} \operatorname{mean}\left( \frac{1}{7} \sum_{d=1}^{7} s_{i,d},; F1_i \right)
\end{equation}
where $s_{i,d}$ denotes the score for dimension $d$ on question $i$, $F1_i$ is the word-level F1 score for the $i$-th question, and $N$ is the total number of evaluated questions. This formulation encourages alignment between LLM-judged quality and factual correctness.

\begin{figure}[h]
\vspace{-2mm}
\centering
\includegraphics[width=0.98\linewidth]{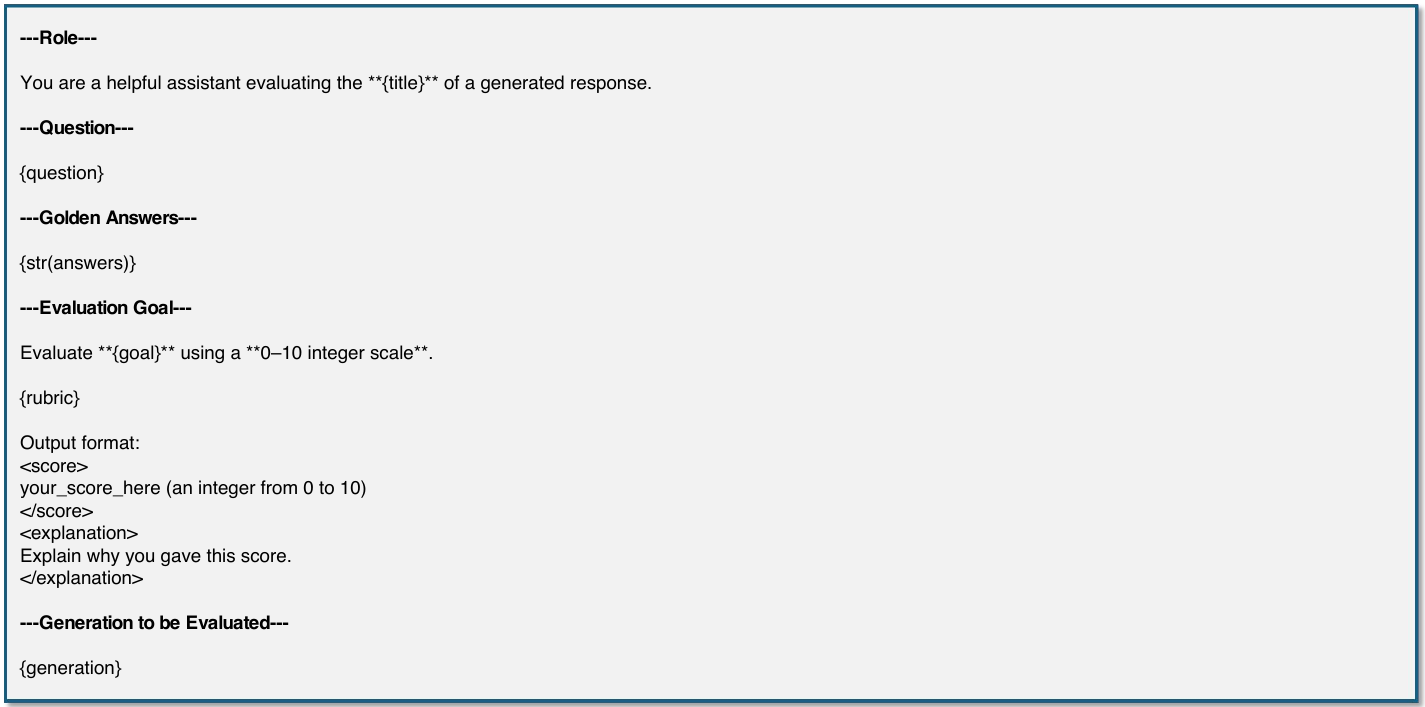}
\caption{\label{prompt4}
Prompt for G-E.}
\vspace{-3mm}
\end{figure}

\begin{figure}[t]
\centering
\includegraphics[width=0.98\linewidth]{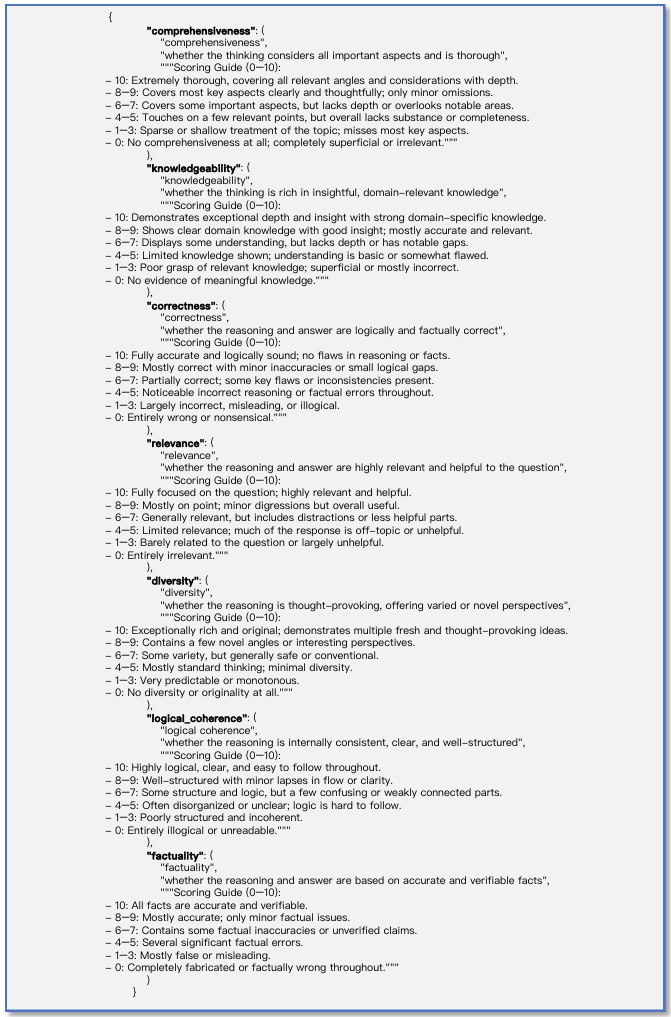}
\caption{\label{prompt5}
Seven Evaluation Dimensions for Generation Quality.}
\end{figure}

\textbf{G-E Prompt.} 
Figure~\ref{prompt4} and Figure~\ref{prompt5} show our generation evaluation prompts. Figure~\ref{prompt4} presents the unified prompt used to score each dimension on a 0–10 scale, while Figure~\ref{prompt5} provides the detailed scoring rubric for all seven dimensions, ensuring consistency and fairness across evaluations.
\clearpage

\section{Baseline Details}

We compare HyperGraphRAG against six representative baselines that cover retrieval-free, chunk-based, and binary graph-based RAG paradigms:

\textbf{NaiveGeneration} is a retrieval-free baseline where the LLM directly answers questions without any external knowledge input. This serves as a lower bound for retrieval-augmented generation.

\textbf{StandardRAG} follows the original RAG design, retrieving top-$k$ text chunks from a flat corpus using dense vector similarity and feeding them into the generator.

\textbf{GraphRAG}~\citep{GraphRAG} constructs a binary relational graph and retrieves community-level summaries linked to query-relevant entities. It uses entity overlap to detect relevant subgraphs.

\textbf{LightRAG}~\citep{LightRAG} enhances retrieval efficiency by using graph indexing and lightweight entity-relation matching over the binary graph, and then combines results with chunk-level retrieval.

\textbf{PathRAG}~\citep{PathRAG} improves graph-based retrieval by selecting paths through the graph that are semantically relevant to the query, using path pruning strategies to reduce redundancy.

\textbf{HippoRAG2}~\citep{HippoRAG2} introduces a high-precision multi-hop retrieval mechanism over binary graphs, using Personalized PageRank to select passage-level nodes connected to the query.

To ensure fairness, all baselines use the same generation prompt (Figure~\ref{prompt3}) and are evaluated under identical conditions, with retrieved knowledge constrained to equivalent token budgets. Each method’s construction and retrieval mechanism is summarized in Table~\ref{comp}.

\section{Hyperparameter Settings}

For all methods, we adopt a unified set of hyperparameters for all models across both the main evaluation in Table~\ref{T1} and the time/cost experiments in Table~\ref{T2} to ensure fair and consistent comparison. For chunk-based methods (e.g., StandardRAG), we retrieve the top-5 chunks using dense similarity. For graph-based methods, including GraphRAG, LightRAG, PathRAG, and HippoRAG2, we retrieve the top-60 relevant elements according to their respective retrieval strategies. HyperGraphRAG performs dual top-60 retrieval over entities and hyperedges, followed by neighborhood expansion. All methods are run using 16 parallel cores and the same generation model (GPT-4o-mini) with temperature 1.0 and a maximum generation length of 32k tokens. Table~\ref{tab:hyperparams} summarizes the detailed hyperparameter configurations used throughout our experiments.

\begin{table}[h]
\centering
\caption{Hyperparameter settings for all methods.}
\small
\label{tab:hyperparams}
\begin{tabular}{lcccc}
\toprule
\textbf{Method} & \textbf{Retrieval Type} & \textbf{Top-$k$ Units} & \textbf{Parallel Cores} & \textbf{Generation Model} \\
\midrule
\rowcolor{gray!10} NaiveGeneration & None & -- & 16 & GPT-4o-mini \\
\rowcolor{gray!10} StandardRAG & Chunk & 5 chunks & 16 & GPT-4o-mini \\
GraphRAG & Entity $\rightarrow$ Community & 60 & 16 & GPT-4o-mini \\
LightRAG & Entity + Relation & 60 & 16 & GPT-4o-mini \\
PathRAG & Graph Path & 60 & 16 & GPT-4o-mini \\
HippoRAG2 & PageRank over Graph & 60 & 16 & GPT-4o-mini \\
\rowcolor{blue!10} \textbf{HyperGraphRAG} (ours) & Entity + Hyperedge & 60 & 16 & GPT-4o-mini \\
\bottomrule
\end{tabular}
\end{table}

\section{Case Study}
To better understand how different methods perform in complex, knowledge-intensive scenarios, we present a case study on the question: \textit{“What type of renal denervation has been shown to demonstrate BP-lowering efficacy over 24 hours in several randomized, sham-controlled trials, particularly for resistant hypertension?”} As shown in Table~\ref{case}, this question requires synthesizing clinical evidence and distinguishing between treatment modalities (e.g., ultrasound vs. catheter-based techniques).

\textbf{HyperGraphRAG} provides the most comprehensive and accurate response, clearly identifying ultrasound renal denervation as the answer, and justifying it with domain-relevant reasoning and structured knowledge. It achieves the highest scores across all dimensions, including \textbf{Correctness (10.00)}, \textbf{Factuality (9.50)}, and \textbf{Knowledgeability (10.00)}, resulting in an overall G-E score of \textbf{92.86}.

\begin{table}[t]
\caption{Case study comparing RAG methods on a complex medical question. The reported metrics reflect the actual scores for this specific example. HyperGraphRAG generates the most accurate and comprehensive answer, achieving the highest scores across all evaluation metrics, while other baselines exhibit factual omissions or ambiguity.}
\vspace{2mm}
\centering
\includegraphics[width=13.8cm]{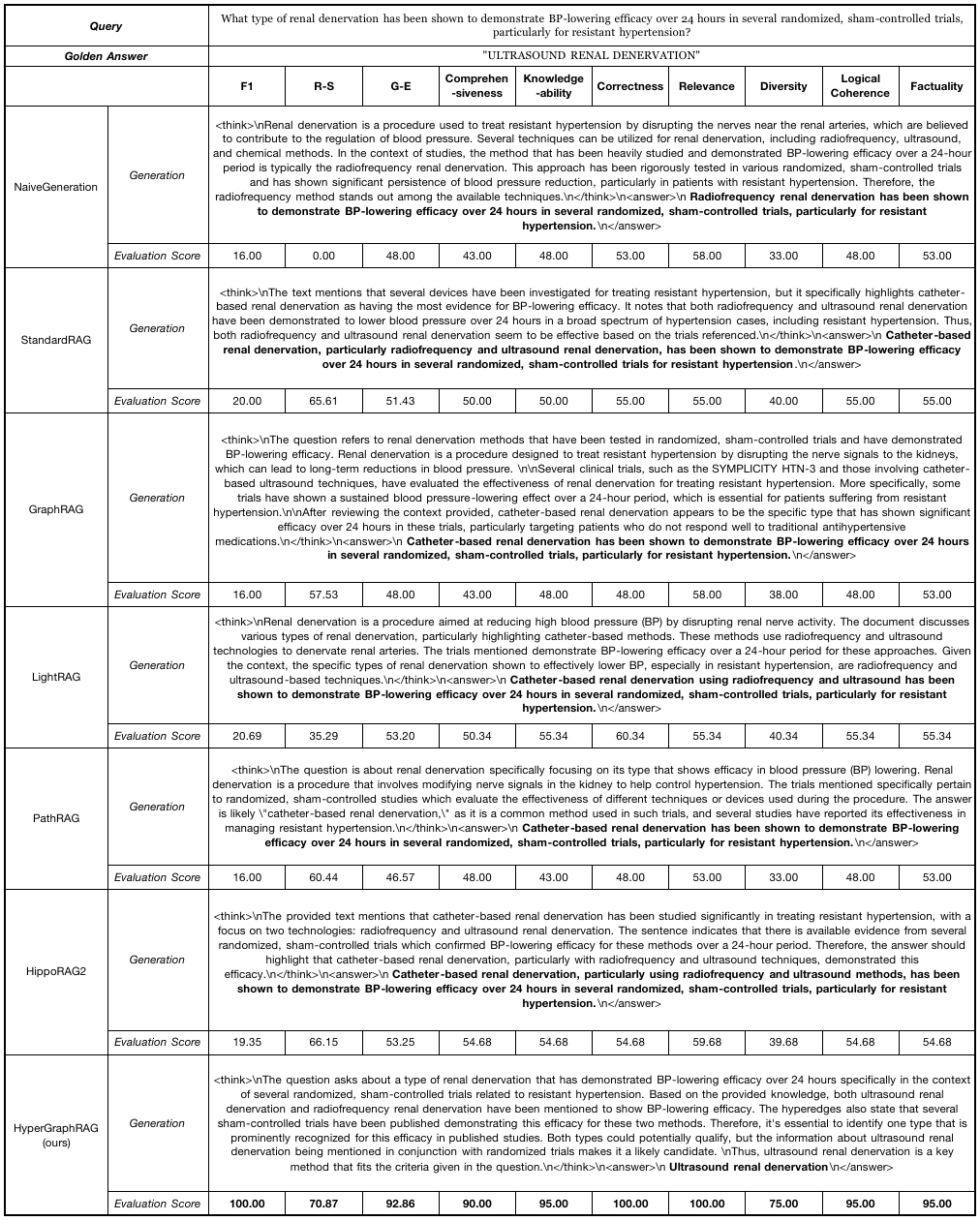}
\label{case}
\vspace{-5mm}
\end{table}

In contrast, baseline methods like \textbf{NaiveGeneration} and \textbf{StandardRAG} produce factually plausible but imprecise or overly generic answers (e.g., mentioning radiofrequency denervation instead), leading to lower scores, particularly in \textbf{Correctness} and \textbf{Factuality}. Graph-based baselines such as \textbf{GraphRAG}, \textbf{LightRAG}, and \textbf{PathRAG} improve coherence but still fall short in domain specificity. Even the best-performing baseline, \textbf{HippoRAG2}, fails to precisely isolate the correct answer, with reduced clarity and comprehensiveness compared to HyperGraphRAG. This case highlights the strength of HyperGraphRAG in integrating multi-entity clinical knowledge through hyperedges, enabling more precise, interpretable, and fact-grounded responses in real-world expert-level tasks.

\section{Limitations and Future Work}
\label{limitations}
\subsection{Multimodal HyperGraphRAG}
While our current framework focuses on textual knowledge, real-world information often spans multiple modalities, including images, tables, and structured metadata. A promising direction is to extend HyperGraphRAG to the multimodal setting by constructing hypergraphs that integrate both textual and non-textual entities (e.g., medical images, diagrams, or structured EHR fields). This would allow the model to reason over complex multimodal relationships, such as “image + report + diagnosis” or “chart + claim + textual guideline,” and enable broader deployment in domains like medicine, science, and law. Future work will explore how to encode, align, and retrieve multimodal hyperedges effectively, while maintaining the structural advantages of hypergraph representations.

\subsection{HyperGraphRAG with Reinforcement Learning}
Another important extension lies in incorporating reinforcement learning (RL) to guide both retrieval and generation. In our current setup, retrieval is driven by fixed similarity metrics, which may not fully capture downstream utility. By formulating hypergraph-based retrieval as a sequential decision-making process, we can apply RL to optimize entity and hyperedge selection policies based on long-term generation rewards—such as factuality, coherence, or user feedback. This would allow HyperGraphRAG to dynamically adapt retrieval strategies to different tasks and domains, leading to more efficient and effective use of structured knowledge.

\subsection{Federated HyperGraphRAG for Privacy-Preserving Retrieval}
Many real-world applications involve sensitive or distributed data that cannot be centralized due to privacy constraints. To address this, we propose to integrate HyperGraphRAG with federated learning techniques, allowing hypergraph construction, retrieval, and generation to occur across decentralized data silos. Each local client can construct its own partial hypergraph and share only anonymized or encrypted embeddings, preserving privacy while contributing to global retrieval. This federated HyperGraphRAG would be particularly beneficial in domains like healthcare or finance, where data sharing is restricted but collective knowledge is crucial for robust decision-making.

\subsection{Toward a Foundation Model for HyperGraph-based Retrieval}

As large language models continue to scale and generalize across domains, a natural extension is to explore the development of a foundation model for HyperGraphRAG. Rather than constructing and retrieving from hypergraphs on a per-task or per-domain basis, we envision a pretrained hypergraph reasoning model that jointly learns representations of entities, relations, and higher-order hyperedges across diverse corpora. This model would encode structural, semantic, and contextual signals in a unified way, and could be adapted to new domains via lightweight fine-tuning. Such a foundation model could also enable transfer learning across knowledge-intensive tasks, reducing the need for domain-specific engineering and improving the sample efficiency of retrieval and generation pipelines. Building this requires scalable hypergraph pretraining objectives, efficient storage formats, and robust generalization strategies, which we leave as future work.

\subsection{Scaling to Harder Tasks and Broader Applications}
Finally, we plan to evaluate HyperGraphRAG on more challenging tasks and diverse real-world applications. This includes settings that require deeper compositional reasoning, such as multi-hop question answering, legal argument generation, or complex scientific synthesis. Additionally, we aim to apply HyperGraphRAG to broader domains beyond the current benchmarks, including policy analysis, education, and open-domain dialogue. These tasks will test the framework’s ability to generalize across domains, handle larger and more diverse knowledge bases, and maintain high-quality generation under increasingly demanding conditions.

\end{document}